\documentclass[a4paper,fleqn]{cas-dc}
\DeclareUnicodeCharacter{202F}{\,}
\DeclareUnicodeCharacter{200B}{}

\usepackage{ulem}
\usepackage{algorithm}
\usepackage{array}
\usepackage{caption}
\usepackage{subcaption}
\usepackage{textcomp}
\usepackage{url}
\usepackage{verbatim}
\usepackage[table]{xcolor}          
\usepackage{graphicx}
\usepackage{amsmath}
\usepackage{tablefootnote}

\usepackage{pifont}

\definecolor{lightgray}{RGB}{255, 255, 255}
\definecolor{darkgray}{RGB}{240, 240, 240}

\usepackage[numbers]{natbib}
\usepackage{microtype} 

\usepackage{flushend}
\usepackage{algorithm}
\usepackage[noend]{algpseudocode}
\usepackage{changepage} 
\makeatletter
\def\BState{\State\hskip-\ALG@thistlm}
\makeatother

\usepackage{array} 

\usepackage{subfiles}
\usepackage{comment}
\usepackage{todonotes}
\usepackage{scalerel}
\usepackage{booktabs}
\usepackage{multirow}
\usepackage{paralist}
\hypersetup{breaklinks=true}  

\def\tsc#1{\csdef{#1}{\textsc{\lowercase{#1}}\xspace}}
\tsc{WGM}
\tsc{QE}

\ExplSyntaxOn
\cs_set:Npn \__first_footerline:
{
  \group_begin:
  \small
  \sffamily
  {\rmfamily\itshape Preprint}
  \group_end:
}
\ExplSyntaxOff

\makeatletter
\let\cas@orig@xfloat\@xfloat
\def\@xfloat#1[#2]{%
  \def\@tempa{#2}%
  \ifx\@tempa\@empty
    \cas@orig@xfloat{#1}[tbp]%
  \else
    \cas@orig@xfloat{#1}[#2]%
  \fi
}
\makeatother
\begin{document}
\setcounter{topnumber}{4}
\setcounter{bottomnumber}{2}
\setcounter{totalnumber}{6}
\renewcommand{\topfraction}{0.85}
\renewcommand{\bottomfraction}{0.5}
\renewcommand{\textfraction}{0.1}
\renewcommand{\floatpagefraction}{0.8}

\shorttitle{3D Reconstruction Techniques in the Manufacturing Domain: Applications, Research Opportunities and Use Cases}    

\shortauthors{Cheng et al.}  

\title [mode = title]{3D Reconstruction Techniques in the Manufacturing Domain: Applications, Research Opportunities and Use Cases}  

\author[1]{Chialoon Cheng}[type=editor,
    orcid=0009-0000-7422-7310]
\cormark[1]
\ead{chialoon@u.nus.edu}

\affiliation[1]{organization={Advanced Robotics Centre},
            addressline={National University of Singapore}, 
            postcode={117608}, 
            country={Singapore}
            }
\affiliation[2]{organization={Independent Researcher},
            }        

\author[2]{Kaijun Liu}
\author[1]{Zhiyang Liu}
\author[1]{Marcelo H Ang Jr}

\cortext[1]{Corresponding author}


\maketitle

\begin{abstract}
This comprehensive review examines the evolution and the current state of the art in three-dimensional (3D) reconstruction techniques in manufacturing applications. The analysis covers both traditional approaches and emerging deep learning methods, showing a critical research gap in unified 3d reconstruction frameworks. Through systematic review of 106 recent publications, we classify reconstruction techniques into three primary categories: data acquisition, point cloud generation, post-processing and applications. Non-contact methods, particularly structured light scanning and stereo vision, have shown significant adoption in manufacturing, with 47\% of surveyed applications focusing on quality inspection. The integration of deep learning has enhanced reconstruction accuracy and processing speed, particularly in feature extraction and matching.  Key applications span design and development (13\%), machining (8\%), process (17\%), assembly (22\%), and quality inspection (40\%). While current technologies achieve sub-millimeter accuracy in controlled environments, challenges persist in handling reflective surfaces and dynamic environments. Our findings indicate a trend toward hybrid systems combining multiple sensor types and processing methods to overcome individual limitations. This survey provides a structured framework for understanding current capabilities and future directions in manufacturing-focused 3D reconstruction. 
\end{abstract}

\section{Introduction}


The 3-Dimensional (3D) Reconstruction has been developed drastically in the past sixty years. The 3D reconstruction in computing industry has always been an important research topic due to its  commercial values. The 3D reconstruction technology uses equipment to acquire object or environment information with 2D image or 3D coordinate data, then analyzes and processes them based on the acquired data. Finally, 3d reconstruction theories and method are applied to reconstruct the contour or texture of the object or environment. Thanks to the current 3D reconstruction technology to be fast, accurate and realistic-like, it has been wide deployed in artificial intelligence, robotics, automated driving, SLAM, Virtual Reality, Digital Twin and many other industries.\newline


3D Reconstruction (3DR) is the process of producing a three-dimensional digital representation of objects or scenes by extracting geometric and texture information from input data, such as images or sensor measurements. Geometric structure is obtained by analyzing multi-view correspondences or depth sensors, while surface texture is mapped to enhance model realism \cite{zhou_comprehensive_2024}. The evolution of 3D reconstruction in manufacturing has been shaped by advances in computer vision, imaging technologies, and artificial intelligence. Over the past decades, its applications have expanded from basic photogrammetry to sophisticated AI-driven neural based reconstruction.\newline

\textbf{1950–1980: Early Foundations}. In the mid-20th century, roberts et al. \cite{robertslg_machine_1963} initiated the research in computer vision, laying the groundwork for modern vision and 3D reconstruction techniques. However, industrial applications remained limited due to computational constraints and the lack of high-resolution imaging technologies. Early developments in structured light and laser-based scanning emerged during this period but were not widely implemented in manufacturing \cite{kiyasu_measurement_1995}.

\textbf{1980–2000: Emergence of Digital Vision Systems}. The 1980s and 1990s saw significant advancements in computer vision and digital imaging. The introduction of structured light scanning \cite{geng_structured-light_2011} and laser triangulation, where the advancing of algorithms enabled more accurate surface reconstruction \cite{shirman_local_1987}, facilitatied applications in industrial metrology and quality control. Concurrently, the integration of 3D scanning with CAD/CAM systems improved design validation and reverse engineering processes \cite{muller_reverse_2000}. The adoption of coordinate measuring machines (CMM) \cite{Mantel1993CoordinateMM} and optical scanning systems\cite{noauthor_nists_2020} in manufacturing marked a shift toward automation.

\textbf{2000–2015: Expansion of Automated 3DR}. With the rise of high-performance computing and industrial automation, 3D reconstruction techniques such as structured light\cite{geng_structured-light_2011}, LiDAR, and multi-view stereo (MVS)\cite{furukawa_accurate_2010} gained prominence. These methods were increasingly applied in defect detection, robotic assembly, and digital twin creation \cite{monostori_cyber-physical_2015}. Machine learning models also began to improve reconstruction accuracy, particularly in manufacturing environments that require precise geometric modeling.

\textbf{2015–2025: AI-Powered and Neural Reconstruction Techniques}. Deep learning has transformed 3D reconstruction in manufacturing over the past decade, with convolutional neural networks (CNNs) \cite{krizhevsky_imagenet_2012} and neural radiance fields (NeRF) \cite{mildenhall_nerf_2020} enabling high-fidelity reconstructions. For example, 3D Gaussian Splatting by Kerbl et al.~\cite{kerbl2023gaussian} represents scenes with parametric 3D Gaussians (with position, covariance, opacity, and view-dependent color via spherical harmonics), enabling high visual quality and real-time rendering (30fps) at 1080p resolution. 
These AI-driven approaches have been integrated with smart factories, improving automation, predictive maintenance, and real-time quality control \cite{slapak_neural_2024}. Additionally, Industry 4.0 technologies, including IoT and cyber-physical systems, have further optimized the efficiency of 3D reconstruction in digital manufacturing workflows \cite{wang_digital-twin_2024}.

From the early use of photogrammetry to the advent of AI-driven neural reconstruction methods, the field of 3D reconstruction has undergone transformative developments that have played a critical role in advancing manufacturing automation and digitalization. As the industry moves toward smarter and more adaptive production systems, future research is expected to focus on real-time, scalable, and high-precision reconstruction techniques that can support the demands of intelligent manufacturing. This survey presents a comprehensive analysis of 3D reconstruction technologies, exploring their underlying methodologies, diverse industrial applications, and prevailing challenges within the context of contemporary manufacturing systems.

\section{Literature search and analysis}
\label{sec:lite_search}

\begin{figure}[!t]
\centering
\begin{minipage}[c]{\linewidth}
  \centering
    \includegraphics[width=\textwidth]{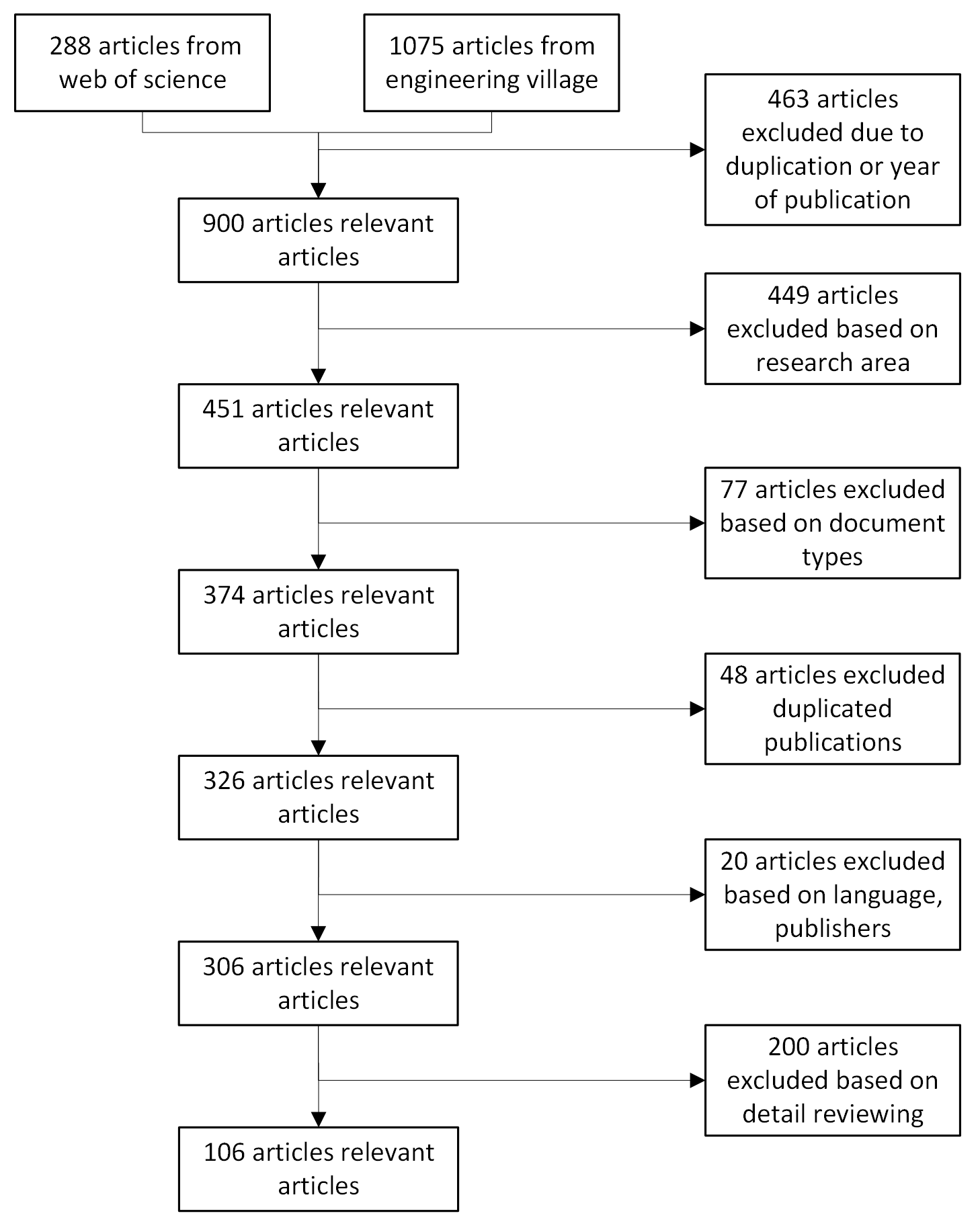}
    \caption{Criteria executed during the search process for relevant literature.}
\label{fig:selection process}
\end{minipage}
\end{figure}

In recent years, numerous studies and reviews have explored 3D reconstruction techniques across various industries. Zhou et al. \cite{zhou_computer_2023} provide a comprehensive overview of computer vision and 3D reconstruction methodologies within the manufacturing sector. Ma et al. \cite{ma_review_2018} review 3D reconstruction techniques in civil engineering, whereas Slapak et al. \cite{slapak_neural_2024} examine the potential applications of neural radiance field approaches in industrial domains. Flores et al. \cite{flores-fuentes_3d_2023} investigate 3D reconstruction techniques applied to specific industries. Although these surveys offer valuable insights, a dedicated review of 3D reconstruction in manufacturing has been absent over the past decade. This study aims to address this gap by comprehensively reviewing methodologies, challenges, and emerging opportunities in the field.

A systematic literature review was conducted, analyzing over 1300 publications in the domain. The selection criteria are summarized in Table \ref{tab:search_exclusion_criteria}, with a visual representation of the selection process provided in Fig. \ref{fig:selection process}. The initial search encompassed publications from 2016 to 2025 across databases such as Engineering Village and Web of Science. Keywords used included “3D Reconstruction AND Manufacturing AND Industry AND (pose estimation OR measurement OR application OR process OR machining OR assembly).”

To maintain focus, studies primarily addressing hardware or sensor technologies were excluded, as this review emphasizes applications spanning data acquisition, processing, and implementation rather than isolated technical components. Additionally, literature outside the manufacturing sector was omitted to ensure relevance. Non-English publications were also excluded due to translation constraints and the potential for misinterpretation.

Applying the selection criteria outlined in Table \ref{tab:search_exclusion_criteria}, the initial query returned approximately 1,300 publications. Further screening based on titles, abstracts, and eligibility criteria refined the dataset to 106 publications directly relevant to 3D reconstruction in the manufacturing industry.

Through a systematic review of 106 recent publications on "3D reconstruction in manufacturing," three critical research gaps have been identified:
\begin{itemize}
    \item The first study classified scanning strategies \newline for manufacturing applications.
    \item No prior research has comprehensively addressed both traditional approaches and machine/deep learning methods for manufacturing applications within a single study.
    \item The first study classified coarse and fine registrations in applications.   
    \item There has been a gap of five years during which relatively less survey papers have discussed or reviewed 3D reconstruction for manufacturing applications.
\end{itemize}

This survey paper is organized into 10 thematic sections. The following provides a brief overview of each section: Section \ref{sec:lite_search} outlines the systematic literature review process, identifying major research themes, methodological trends, and gaps in 3D reconstruction studies. Section \ref{sec:overview_recon_framework} introduces a unified 3D reconstruction framework, detailing core pipeline stages from data acquisition to application in industrial and robotic contexts. Section \ref{sec:3d_representation} explores various 3D data representations, evaluating their strengths and limitations for specific manufacturing, inspection, and assembly-related applications. Section \ref{sec:data_acquisition} discusses data acquisition technologies, including contact or non-contact based sensors, configurations, and scanning strategies used to capture accurate 3D information in industrial environments. This section discusses the growing trend of non-contact sensors becoming the dominant technology in 3D reconstruction. Compared to traditional contact-based methods, non-contact sensing devices—such as structured light, laser scanners, and time-of-flight cameras—offer faster data acquisition, reduced mechanical complexity, and the ability to capture detailed surface geometry without physical interaction. These advantages have led to their widespread adoption in manufacturing, robotics, and quality inspection. As the demand for high-resolution, real-time 3D data continues to rise, non-contact sensors are increasingly replacing conventional systems, shaping the future of digital reconstruction technologies. Section \ref{sec:traditional_technique} presents traditional 3D reconstruction methods like stereo vision and structured light, emphasizing their theoretical foundations and practical performance limitations. Section \ref{sec:ml_dl_approaches} examines recent machine learning and deep learning approaches, such as NeRF, Gaussian Splatting
and CNN-based models, for improving reconstruction robustness and scalability. Section \ref{sec:applications} showcases real-world applications of 3D reconstruction in automotive manufacturing, including digital twins, quality inspection, and robotic path planning. Section \ref{sec:challenges_and_future} addresses current technical challenges and anticipates future opportunities for expanding 3D reconstruction in dynamic, complex manufacturing environments. Section \ref{sec:conclusion} concludes the paper by synthesizing key findings and proposing future research directions for advancing industrial 3D reconstruction technologies.

\begin{table}[h]
    \centering
    \caption{Search and Exclusion Criteria}
    \label{tab:search_exclusion_criteria}
    
    \resizebox{\columnwidth}{!}{
    \begin{tabular}{@{}ll@{}}
        \hline
        \textbf{Search Criteria} & \textbf{Description} \\
        \hline
        Search keywords & 3D Reconstruction AND Manufacturing AND Industry AND \\
                        & (pose estimation OR measurement OR assembly) \\
        Period of publications & 2016--2025 \\
        Publication type & Journal articles or conference articles \\
        Research Areas & Computer vision, robotics, quality inspection, assembly \\ \hline
    \end{tabular}
    }
    
\end{table}

\section{Overview of 3D reconstruction framework}
\label{sec:overview_recon_framework}

\begin{figure*}[!t]
	\centering
	\includegraphics[width=0.9\textwidth]
    {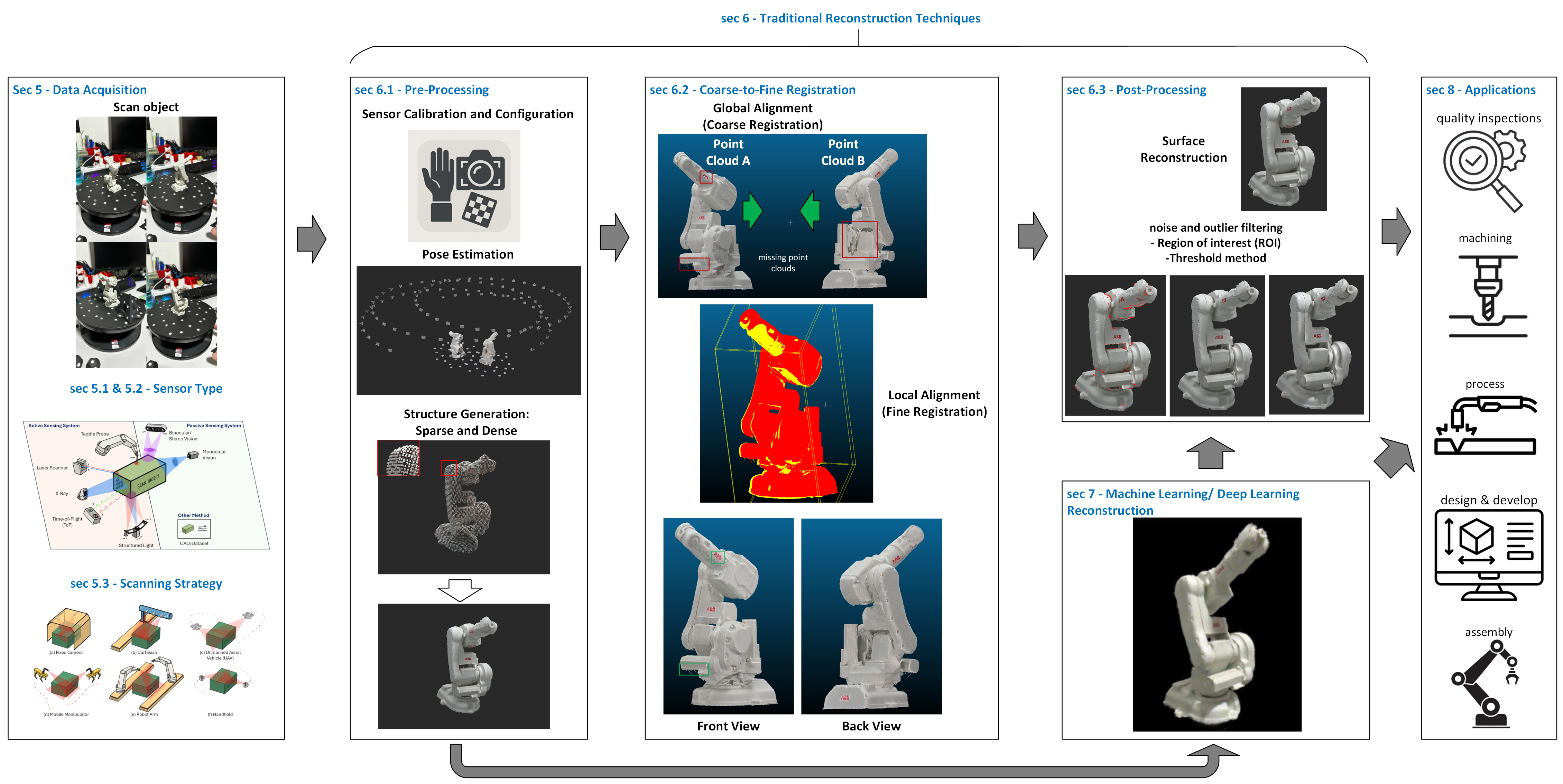}
    	\caption{An overview of 3D Reconstruction pipeline.
    }
	\label{fig:3drecon_overview}
\end{figure*}

The 3D reconstruction framework, illustrated in Fig. \ref{fig:3drecon_overview}, starts with data acquisition and has become integral to intelligent manufacturing systems by enabling the creation of digital models from physical components for tasks such as inspection, assembly planning, reverse engineering, and process control. Growing demands for automation, precision, and real-time responsiveness have propelled 3D reconstruction methods beyond experimental stages toward scalable frameworks suited for industrial environments. Such frameworks integrate diverse components—from data acquisition and preprocessing to registration, representation, and interpretation—each adapted to the geometric complexity, accuracy, and throughput demands of specific manufacturing applications.

The primary goal of this framework is to convert raw sensor inputs into metrically precise and semantically rich digital representations that facilitate decision-making and robotic operations. Data acquisition employs various sensors—including structured light, LiDAR, stereo cameras, laser scanners, and CT imaging \cite{thomas_history_2013}—selected based on task-specific requirements such as resolution, material properties, and surface reflectivity. Sensor choice and scanning approach—whether fixed, robot-mounted, or mobile—critically influence data quality, acquisition speed, and system flexibility.

Following acquisition, preprocessing modules handle sensor calibration, noise removal, and feature extraction to ensure geometric consistency and prepare the data for reconstruction. These steps are critical for aligning disparate datasets collected under varying poses or lighting conditions. Multi-view data often requires robust registration pipelines, typically divided into coarse (global) and fine (local) alignment stages. Techniques such as RANSAC \cite{fischler_random_1981}, FPFH matching \cite{rusu_fast_2009}, and Iterative Closest Point (ICP) \cite{besl_method_1992} are widely used in this stage to achieve accurate fusion of multiple scans into a coherent coordinate system.

The resulting dense point cloud undergoes refinement through post-processing steps like downsampling, meshing, or implicit surface fitting, converting raw geometry into formats suitable for downstream tasks including CAD comparison, tolerance verification, and robotic path planning. High fidelity and timely feedback are critical in manufacturing operations such as in-line inspection and assembly verification. Accordingly, the framework supports both explicit geometric representations (e.g., meshes, point clouds) and implicit ones. Signed Distance Functions (SDFs) encode each 3D point by its signed distance to the nearest surface, providing smooth implicit geometry \cite{park2019deepsdf}. Truncated SDFs (TSDFs) clamp this distance within a fixed band, enabling robust multi-view fusion and real-time reconstruction in systems like KinectFusion \cite{curless1996volumetric,newcombe2011kinectfusion}.

Recent advances incorporate deep learning to enhance feature extraction, pose estimation, and shape inference from incomplete or noisy data, improving robustness under challenging conditions. However, these techniques demand substantial data, generalization capabilities, and interpretability.

This study proposes a comprehensive 3D reconstruction framework for manufacturing, synthesizing insights from over 106 recent studies. It systematically explores each pipeline stage—from sensor selection and scanning methodologies to both classical and learning-based reconstruction techniques—with a focus on application relevance, scalability, and accuracy to support ongoing progress in smart manufacturing systems.

\section{3D Representation in applications}
\label{sec:3d_representation}
This section provides a comparative analysis of prominent 3D representations—such as SDF, TSDF, NeRF, Gaussian Splatting, point clouds, meshes, voxels, Boundary Representation (B-Rep) and Constructive Solid Geometry (CSG) in table \ref{tab:3d_representation_summary_with_process}. Each representation is evaluated in terms of its core characteristics, advantages, limitations, and specific applications in manufacturing tasks including inspection, assembly, and design optimization.

The \textbf{Signed Distance Field (SDF)} is a volumetric function that assigns to each point in space the signed distance to the nearest surface, where negative values indicate interior regions and positive values indicate exterior \cite{park2019deepsdf}.
The surface itself is recovered as the zero-level set of this function, i.e., the set of points where the signed distance equals zero \cite{osher1988fronts}.
This representation is particularly advantageous in design and development processes, where smooth and differentiable surface modeling is required. SDFs support GPU-based parallelization and are well-suited for applications such as precision measurement, surface reconstruction, and simulation-driven design optimization. In the manufacturing domain, SDFs have been applied to enhance the accuracy of 3D surface reconstructions, which is critical for dimensional verification and tolerancing in automotive body components. As shown in \cite{hu_point_2024,chen_implicit_2023}, SDFs facilitate digital prototyping by enabling robust gradient-based operations, such as local surface deformation and topology optimization. Despite its advantages, the SDF requires high-resolution volumetric grids, resulting in elevated memory consumption and the necessity for meshing (e.g., via Marching Cubes\cite{Lorensen1987Marching}) for visual or physical interpretation. The Truncated Signed Distance Field (TSDF) is a memory-efficient variant of the SDF that truncates distance values beyond a certain threshold to enable real-time volumetric fusion. TSDF is predominantly utilized in robotic assembly and automated inspection pipelines, where fast and incremental updates to the scene are required. Its application includes transparent object detection, grasp planning, and pose refinement during assembly tasks, especially in environments with partial occlusions and motion constraints. As discussed in \cite{dai_graspnerf_2023,liu_rgbgrasp_2023,breyer_volumetric_2021}, TSDFs enable simultaneous localization and mapping (SLAM) in manufacturing cells, allowing robots to perceive their surroundings with updated 3D maps for interaction. However, TSDF suffers from reduced geometric fidelity at object boundaries and becomes memory-intensive in large-scale scenes, limiting its scalability without hierarchical or sparse voxel structures.

\textbf{Neural Radiance Field (NeRF)} 
represent a scene as a continuous function parameterized by an MLP that maps 3D coordinates and viewing directions to color and density, enabling novel view synthesis from sparse 2D images \cite{mildenhall_nerf_2020}
and it has emerged as a powerful tool for visual inspection and virtual simulation in manufacturing. It excels in reconstructing challenging surfaces such as specular, reflective, or occluded materials where traditional methods fail. For instance, \cite{zhang_neural_2023} applied NeRF to create realistic digital twins of manufactured parts for operator training and automated quality control. Similarly, \cite{ichnowski_dex-nerf_2021} demonstrated its effectiveness in transparent object inspection for semiconductor production lines. However, industrial adoption faces challenges due to long training times and lack of explicit geometry for metrological analysis.

\textbf{Gaussian Splatting} 
models a scene as a set of 3D Gaussians that are projected and rasterized in screen space, enabling real-time photorealistic rendering from multi-view observations\cite{kerbl2023gaussian} and it bridges explicit geometry and neural rendering for real-time manufacturing applications. Its ability to aggregate multi-view observations makes it ideal for rapid defect detection in occluded areas, providing immediate visual feedback in smart inspection cells \cite{ito_free-viewpoint_2024}. While proven effective for photorealistic part visualization in quality control, its reliance on view consistency rather than physical constraints limits deployment in heavy-duty assembly tasks requiring force interaction modeling.

\textbf{Point clouds} representation has emerged as a fundamental data structure in manufacturing and computer vision, offering a detailed yet flexible means of capturing three-dimensional geometry. By encoding spatial coordinates of sampled points, it enables accurate modeling of complex surfaces, thereby supporting a variety of analysis and automation tasks \cite{liu_deep_2019}.In general, SFM and photogrammetry-generated point clouds usually have a low and sparse point density, while 3D scanners, LiDAR, and depth sensors can generate point clouds with more points \cite{liu_deep_2019}. They serve as the backbone of numerous applications, including dimensional inspection, reverse engineering, and virtual assembly. In \cite{pas_grasp_2017}, dense segmented point clouds were used to reconstruct the digital twin of a manufacturing environment, enabling efficient segmentation and object labeling for downstream automation. Despite their versatility, point clouds lack connectivity and surface continuity, often requiring meshing or fitting for downstream analysis. Their unstructured nature also introduces challenges in noise filtering and registration.

\textbf{Meshes}, composed of vertices and faces, provide a topologically structured representation ideal for CAD-based comparisons, finite element analysis, and tolerance verification \cite{Homri2017Tolerance}. In manufacturing, meshes are frequently used to validate scanned parts against nominal CAD models to identify deviations or deformations. As indicated in \cite{xie_aircraft_2020}, mesh-based representations are employed for virtual assembly simulation and collision detection. Their major advantage lies in their compatibility with simulation and metrology tools; however, generating high-quality meshes from noisy point clouds remains computationally intensive and prone to error in complex scenes.

\textbf{Voxel grids} divide 3D space into tiny boxes, called voxels, that form a regular grid \cite{Xu2021Voxel}. Often used in early-stage occupancy mapping, collision detection, and robot motion planning. Their uniform structure supports GPU acceleration and is suitable for training data-driven models. In \cite{ito_free-viewpoint_2024}, voxel grids were used to model workspace occupancy for robotic arms performing automated tightening and part insertion. Although voxels are easy to process, they suffer from resolution limits and high memory usage, especially for large-scale scenes.


\textbf{Boundary representation (B-rep)} models 3D solids by explicitly defining their boundaries—faces, edges, and vertices—along with topological relationships\cite{Mantyla1988Solid}. This representation enables precise geometric and topological modeling widely used in CAD for design, simulation, and manufacturing \cite{liu_manufacturing_2025}. Liu et al. \cite{liu_advancing_2024} focused on mesh generation on composite surfaces with boundary representation, handling and repairing mesh boundaries, discrete and continuous surfaces, and virtual topology merging on B-Rep constructs. Resulting in higher inspection accuracy.

\textbf{Constructive Solid Geometry (CSG)} represents solids by combining basic primitives, such as cubes, cylinders, and spheres, through Boolean operations (union, intersection, difference)\cite{Laidlaw1986CSG}. Rather than storing explicit boundaries, CSG forms complex objects via a Boolean expression tree of primitives. Lv et al. \cite{lv_teapot_2018} demonstrated its effectiveness in producing valid solid models, enabling efficient 3D reconstruction and supporting manufacturing tasks like teapot prototyping.

\begin{figure}[!t]
\centering

    \includegraphics[width=0.48\textwidth]
    {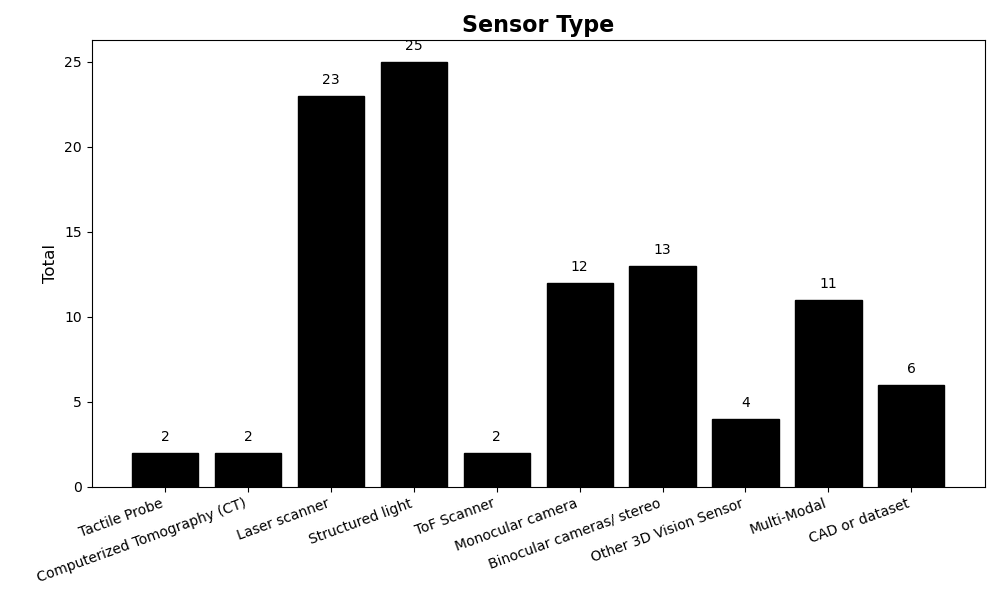}
    \caption{Different sensor types based on population from survey papers}
\label{fig:SENSORTYPE}
\end{figure}

In conclusion, this section examined 3D representations in the context of their targeted applications, the subsequent section will address data acquisition. This includes a discussion of the types of sensors (Fig. \ref{fig:SENSORTYPE}) employed and the scanning strategies adopted to achieve the desired accuracy and sensing range.

\begin{table*}[htbp]
\centering
\caption{Summary of 3D Representations: benefits, drawbacks, process roles, and manufacturing applications.}
\label{tab:3d_representation_summary_with_process}

\resizebox{0.9\textwidth}{!}{%
\begin{tabular}{p{1.5cm} p{3.2cm} p{4.2cm} p{4.2cm} p{3.0cm} p{6.2cm} p{3.7cm}}
\hline
\textbf{Representation} & \textbf{Type} & \textbf{Key Benefits} & \textbf{Key Drawbacks} & \textbf{Process} & \textbf{Application} & \textbf{Papers} \\
\hline

Implicit & Signed Distance Field (SDF)\cite{newcombe2011kinectfusion}/ Truncated SDF (TSDF)\cite{curless1996volumetric} & Smooth, continuous surface representation. Suitable for differentiable rendering and shape optimization. GPU-friendly parallel computation. & Requires high-resolution grids → high memory. Needs meshing (e.g., marching cubes\cite{Lorensen1987Marching}) for visualization or use. & Design and Develop & Enhanced 3D surface reconstruction precision measurement & \cite{hu_point_2024,chen_implicit_2023} \\
 &  &  &  & Quality Inspection & Sub-surface defect mapping & \cite{ding_robust_2019} \\
 & & Efficient for real-time 3D fusion (SLAM). Captures local surface proximity well. & Memory intensive for large scenes. Reduced detail due to truncation. & Assembly & Transparent object detection grasp planning cluttered environment handling & \cite{dai_graspnerf_2023,liu_rgbgrasp_2023,breyer_volumetric_2021} \\ \hline
 & Neural Radiance Field (NeRF)\cite{mildenhall_nerf_2020} & Realistic rendering under complex lighting. Captures view-consistent textures. & Long training/rendering time. No explicit geometry output. & Quality Inspection & Photorealistic modeling of products for visual inspection & \cite{zhang_neural_2023}\\
 &  &  &  & Process & Handling specular/transparent objects &\cite{ichnowski_dex-nerf_2021}\\ \hline
 & Voxel Grid\cite{Xu2021Voxel} & Regular grid structure supports simulation. Compatible with 3D CNNs and occupancy modeling. & Coarse resolution or high memory cost. Less precise than mesh/B-Rep. & Design and Develop & Collision field modeling. &\cite{harabin_deep_2023}\\
 &  &  &  & Quality Inspection & Inspecting assembled seal integrity in 3D internal defect detection in large components &\cite{da_silva_santos_3d_2023,le_reun_improving_2024}\\
 &  &  &  & Process & Voxelized grasp planning in dense clutter environment & \cite{pas_grasp_2017} \\
\hline\hline
Explicit & Dense Point Cloud\cite{liu_deep_2019} & High-resolution, sensor-direct representation. Flexible for mesh conversion or registration. & No surface connectivity. Large data size, noisy if unfiltered. & Design and Develop & Scan-CAD comparison for reverse engineering extracting surface features to aid virtual assembly design improving mesh reconstruction quality from scan data scanned point data directly for CFD simulations & \cite{zhang_cad-aided_2022,ma_application_2019,mu_structural_2024,jaiswal_mesh-driven_2024}\\
 &  &  &  & Quality Inspection & Surface defect detection (e.g. scratches, dents, porosity) on parts manufacturing tolerance checks &\cite{li_novel_2023,mosca_ransac-based_2020,wang_density-invariant_2021,xie_aircraft_2020,huang_rapid_2022,fu_3d_2023,zhou_aviation_2024,liu_high-accuracy_2022,bauer_registration_2021,zhang_screening_2023,ding_three-dimensional_2020,xu_3d_2022}\\
 &  &  &  & Assembly & Robot path planning and collision avoidance using detailed environment scans for assembly purpose guiding human assembly via 3D scanned model overlays &\cite{jiang_geometry_2024,nigro_assembly_2023,nguyen_revolutionizing_2024,wang_digital-twin_2024,lu_cfvs_2023,li_rearrangement_2023,chang_vision-based_2021,wang_pose_2020,tan_multi-robot_2023,skeik_6d_2022,sileo_vision-enhanced_2024,theodoropoulos_cyber-physical_2023,xu_fast_2017}\\
 &  &  &  & Machining & Detecting machining errors or surface deviations on workpieces &\cite{xi_deep_2024}\\
 &  &  &  & Process & Robotic cell path planning and obstacle avoidance for process purpose creating digital twins of factory scenes from point cloud data for layout and scene understanding &\cite{zhou_learning-based_2022,gong_point_2017,chang_eye--hand_2016,li_c-lfnet_2024,zhang_3d_2019,peng_tool_2024,zheng_semantic_2025}\\ \hline
 & Sparse Point Cloud\cite{liu_deep_2019} & Fast, low-memory representation. Suitable for rapid or partial scans. & Lower detail → unsuitable for high-precision tasks. Incomplete surface data. & Quality Inspection & Quick, localized scanning for error analysis on parts simplifying a dense cloud to a sparse set to facilitate inspection of very complex assemblies high-speed hole geometry tolerancing by comparing sparse points to CAD targeted measurement point acquisition for large structures (e.g. ship hull blocks) &\cite{gai_construction_2020,navarro-jimenez_reconstruction_2023,jin_automatic_2016,zhao_position_2019,kim_highly_2024}\\
 &  &  &  & Machining & Extracting key features (hole centers, edges) from a dense cloud into a sparse set for machining guidance &\cite{zheng_laser_2025}\\
 &  &  &  & Process & augmented reality tracking of assembly progress using a sparse model for worker training and guidance Measuring and checking dimensions (extract center points) &\cite{xie_measurement_2024,paulo_lima_markerless_2017} \\ \hline
 & 3D Mesh (Polygonal Surface)\cite{Homri2017Tolerance} & Continuous surface model, widely supported. Efficient rendering and usable in CAD/CAM. & Requires cleanup when generated from scans. Approximate—may lose fine detail. & Design and Develop & Reverse engineering: converting point cloud data into accurate mesh/CAD models repairing or modifying legacy designs by mesh editing and then updating the CAD building digital twins of environments or products for design context and scene understanding &\cite{ni_three-dimension_2018,yu_rapid_2019,m_minos-stensrud_towards_2018}\\ 
 &  &  &  & Quality Inspection & Dimensional and tolerance inspection using meshes as reference surfaces assessing part damage or deformation by comparing mesh to nominal  using multi-sensor data for detecting defects  &
\cite{yang_dpps_2023,iuso_cad-based_2021,zhao_three-dimensional_2021,kinnell_autonomous_2017,zeng_registration_2021,sentenac_automated_2018,riffo_active_2022}
 \\
 &  &  &  & Assembly & Robot path planning and collision avoidance &\cite{wang_robot_2024}\\
 &  &  &  & Machining & Tool path planning on mesh models of parts &\cite{wang_trajectory_2022}\\
 &  &  &  &  & Process: Tool path planning on mesh models of parts &\cite{prezas_multi-purpose_2024,zhang_robot_2019}\\ \hline
 & Depth Map (2.5D Image)\cite{Qi2018GeoNet} & Lightweight, 2.5D view with visual clarity. Simple to extract from depth sensors. & Single viewpoint only. No full geometry or occluded area. & Quality Inspection & Depth image analysis Error heat maps used to visualize machining or fabrication deviations & \cite{kumar_surface_2018,chew_-process_2024}\\ \hline
 & Boundary Representation (B-Rep)\cite{Mantyla1988Solid} & Exact geometry (e.g., NURBS\cite{Piegl1995NURBS}). Parametric, supports CAD operations and simulations. & Hard to derive from raw scan. Complex to render or convert. & Design and Develop & update models of free-form parts​ of reverse-engineered CAD  Generating cleaner CAD surfaces from noisy scans Reverse designing and remaking parts using B-spline surface fitting and 3D printing for legacy parts Improving efficiency and reducing cost; using B-Rep assemblies (STEP format) combined with virtual assembly to check tolerances and interferences before manufacturing & \cite{yan_point_2021,falheiro_cad_2021,gao_3d_2018,wang_application_2022}
 \\ 
 &  &  &  & Assembly & Simulation with STEP models & \cite{wang_assembly_2021} \\
 &  &  &  & Process & Inline inspection of reverse-engineered models during manufacturing processes & \cite{falheiro_cad_2021} \\ \hline
 & Constructive Solid Geometry (CSG)\cite{Laidlaw1986CSG} & creates complex solids by combining simple geometric primitives with Boolean operations like union, intersection, and difference. & Lacks explicit faces, edges, or vertices, making detailed feature access and boundary extraction difficult. & Design and Develop & applies reverse engineering and rapid prototyping for reconstructing and fabricating a 3D model. & \cite{lv_teapot_2018,andujar_solid_2022}
\\ \hline
 & 3D Gaussians (Gaussian Splatting)\cite{kerbl2023gaussian} & Real-time, smooth rendering. Highlights shape via transparency & Low geometric accuracy. Limited toolchain support. & Quality Inspection & Rapidly visualize scan data using Gaussian splats to reveal hidden defects by merging multiple viewpoints &\cite{ito_free-viewpoint_2024}\\
\hline
\end{tabular}%
}

\end{table*}

\section{Data Acquisition}
\label{sec:data_acquisition}

\begin{figure*}[!t]
	\centering
	\includegraphics[width=0.9\textwidth]
    {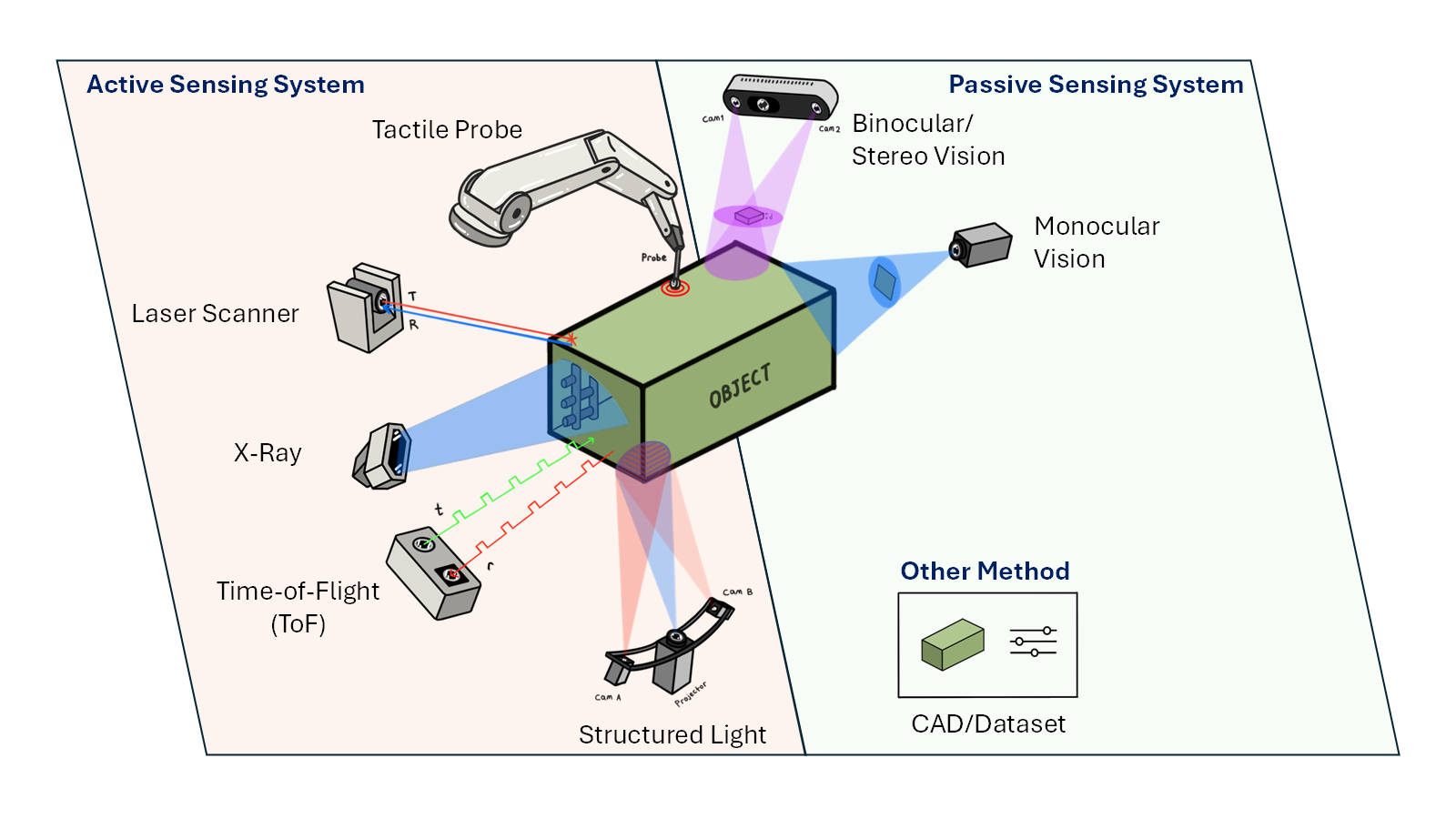}
    	\caption{The sensor types and their data acquisition methods are categorized into active and passive systems. The left section represents active systems, while the right section represents passive systems. Active systems include tactile probes, laser scanners, X-ray, Time-of-Flight (ToF), and structured light. Passive systems include monocular and binocular vision.}
	\label{fig:sensor_overview}
\end{figure*}

In 1997, Várady et al. \cite{varady_reverse_1997} classified 3D reconstruction data acquisition methods into contact and non-contact sensing approaches.\newline

\textbf{Contact-Based Sensing}, commonly referred to as tactile sensing, entails the direct physical interaction with a target object to acquire three-dimensional point cloud data \cite{intwala_review_2016}. This approach typically employs motion-enabled devices such as Coordinate Measuring Machines (CMMs) or robotic manipulators, wherein a measurement \textit{probe} is affixed to a multi-degree-of-freedom actuation platform, as illustrated in Fig. \ref{fig:sensor_overview}. The number of degrees of freedom is inherently dictated by the specific type of motion system utilized. One of the principal advantages of contact-based sensing lies in its exceptional measurement accuracy, rendering it particularly effective for applications requiring high-precision, small-scale metrology. As a result, CMMs have remained the industry benchmark for dimensional inspection across various manufacturing domains. In automotive body inspection and other large-scale use cases, these systems are capable of achieving measurement accuracies on the order of 0.1 mm.

Despite these strengths, contact-based techniques exhibit several inherent limitations. Their constrained measurement volume, difficulty in accommodating large or geometrically complex parts, and the necessity for probe radius compensation introduce operational challenges. Moreover, the physical interaction with the surface can lead to inaccuracies, particularly when inspecting deformable materials, and may pose a risk of surface damage. In addition, the inherently sequential nature of the measurement process contributes to slower data acquisition rates, thereby reducing overall throughput.\newline

\textbf{Non-Contact based Sensing} has gained significant traction in industrial environments due to its ability to perform three-dimensional reconstruction without requiring physical contact with the target object. These techniques infer the spatial geometry of objects by analyzing externally captured data, thereby eliminating mechanical interference during measurement. According to the classification introduced by Isgro et al. \cite{isgro_open_2005}, non-contact methods can be broadly divided into active and passive approaches.\newline

Active methods operate by emitting energy—such as structured light patterns or acoustic pulses—toward the object surface and analyzing the reflected or scattered signal to reconstruct depth information. In some cases, sensor parameters, such as focal length or exposure, may be dynamically modulated to enhance measurement fidelity. Conversely, passive techniques do not emit additional energy but instead rely on ambient illumination. They reconstruct the 3D structure by interpreting naturally occurring light reflections or shadows captured across multiple viewpoints.

\subsection{Active Sensing Systems}
\textbf{Active sensing systems} include techniques such as Computerized Tomography (CT), Laser Scanning \cite{cartwright_3d_2008}, Structured Light, Time-of-Flight (ToF) \cite{lewellen_time--flight_1998}, and RGB-D Imaging \cite{pollefeys_detailed_2008}. These methods leverage optical instruments to capture surface data, which is then processed to reconstruct the 3D geometry of the object.\newline

\textbf{Computerized Tomography (CT)} - Non-optical sensing techniques include ultrasound \cite{antonio_aceves-fernandez_survey_2019} and Industrial Computed Tomography (CT) \cite{thomas_history_2013}. Industrial CT scanning employs X-ray irradiation to generate 3D representations of scanned objects. The system comprises a transmitter emitting electromagnetic radiation, receivers capturing the signal, and a control unit that filters, amplifies, and transmits the data for processing. The primary advantage of electromagnetic tracking is real-time 3D motion capture, although it is limited by magnetic field constraints and restricted sensor movement. A key application of CT is in industrial inspection, where Le Reun et al. \cite{le_reun_improving_2024} utilized X-ray CT for high-precision reconstruction of large aircraft components.\newline

\textbf{Laser Scanning} measures real-world objects by emitting a laser beam onto a surface and analyzing the reflected signal. Elgazzar et al. \cite{elgazzar_active_1998} introduced triangulation-based laser scanning, a cost-effective method that improves measurement accuracy for indoor reconstruction. Boehler et al. \cite{boehler_investigating_2003} further analyzed the performance of different 3D laser scanners, demonstrating sub-millimeter precision suitable for quality inspection and reverse engineering \cite{blais_review_2004}. However, laser scanners encounter difficulties with highly reflective, transparent, or dark surfaces, often requiring additional preprocessing to mitigate data loss \cite{lichti_effects_2002}.\newline

\textbf{Time-of-Flight (ToF)} technology operates by emitting pulsed infrared light  and measuring the time delay before its return to receiver to calculate depth information (shown in Fig. \ref{fig:sensor_overview}). ToF cameras facilitate real-time 3D point cloud acquisition, offering advantages such as rapid data capture compared to structured light and stereo vision \cite{hansard_time--flight_2013}. However, ToF sensors are sensitive to motion, leading to depth estimation inaccuracies when objects move within the scene \cite{tan_cloud-based_2012}.\newline

\textbf{Structured Light} is widely utilized in 3D reconstruction \cite{fofi_comparative_2004, wang_generation_2021, geng_structured-light_2011}. The system typically consists of a projector, cameras, and a calibration reference. Structured patterns are projected from the \textit{projector} device onto the object, and the deformations in these patterns are captured by cameras (\textit{cam A} and \textit{cam B}) in Fig. \ref{fig:sensor_overview}. Depth information is then computed using triangulation-based methods \cite{otoole_3d_2014, brown_overview_2000}. Structured light can be classified into point-based, line-based, surface-based, network-based, and color-coded approaches \cite{geng_structured-light_2011}. However, movement-induced distortions in the projected patterns can affect accuracy \cite{tan_cloud-based_2012}.

Structured light techniques are not monolithic and include several specialized modalities. Fringe projection profilometry (FPP) is a mainstream structured-light approach for 3D reconstruction of diffuse surfaces, known for its high accuracy and resolution in close-range measurements. In FPP, a series of phase-shifted sinusoidal fringe patterns is projected onto the object, and the distorted fringe images captured by the camera are analyzed to retrieve per-pixel phase values that correspond to object surface heights. This phase-encoding technique is robust against noise and provides dense, pointwise depth information, since each pixel’s intensity sequence encodes a unique phase value \cite{li_single-shot_2024,tan_robust_2021}. The result is a high-resolution 3D point cloud or mesh of the object surface. However, because the raw phase is wrapped modulo $2\pi$, FPP requires phase-unwrapping algorithms (spatial or temporal) and typically multiple projected patterns to resolve ambiguities, which can restrict its use to static or slow-moving targets\cite{li_single-shot_2024}.

Fourier transform profilometry (FTP) is a variant of structured-light fringe projection that enables one-shot 3D shape capture using a single fringe image. Instead of multiple phase-shift patterns, FTP projects a single sinusoidal fringe pattern of known spatial frequency onto the object and records the deformed pattern with a camera. The captured image is then processed in the frequency domain: by applying a 2D Fourier transform and band-pass filtering, the fundamental frequency component (carrier) of the fringe is isolated, and an inverse Fourier transform yields the wrapped phase map of the object surface from that single frame\cite{su_fourier_2001}. The primary advantage of this method is its ability to compute shape from just one image, making it well-suited for dynamic scenes or real-time measurements where the object or camera may not be stationary. FTP thereby eliminates the motion artifacts associated with multi-pattern techniques; however, it demands a high signal-to-noise ratio and careful calibration since any image noise or extraneous frequencies can affect the phase extraction. In practice, Fourier profilometry has been successfully applied in scenarios like vibration measurement and rapid prototyping inspection, where its single-shot nature provides 3D data at video rates, albeit often with slightly lower precision than multi-shot phase-shifting methods under the same conditions \cite{su_fourier_2001}.

Beyond structured light, laser scanning, and ToF technologies, several high-precision interferometric methods have seen industrial adoption. White light interferometry (WLI) is a non-contact optical technique offering nanometer-scale vertical resolution for 3D surface topography measurement. Using a broadband light source, it identifies peak fringe contrast through vertical scanning to reconstruct height profiles with sub-nanometer precision. Widely adopted in industrial metrology, WLI provides high data density and traceability, ideal for characterizing machined surfaces \cite{wu_research_2023}. Commercial systems like Zygo NewView and Bruker Contour achieve excellent repeatability. While slower over large areas due to sequential scanning, advances in high-speed processing and compact sensor integration have enabled real-time, in-situ surface monitoring for precision manufacturing and closed-loop quality control \cite{wu_research_2023}. Similarly, Digital holographic interferometry is a fast, high-precision 3D measurement technique used in manufacturing. It captures full-field surface data with sub-micron accuracy by reconstructing light wavefronts reflected from objects. Multi-wavelength systems can measure both reflective and matte surfaces, making them ideal for inline inspection of parts like automotive components and microelectronics. Advanced algorithms allow real-time imaging and refocusing, overcoming challenges with depth, focus, and surface complexity in industrial settings\cite{LAM2021010002}.

Active sensing system are well-suited for high-precision 3D reconstruction. They offer high depth accuracy—often at centimeter or sub-millimeter scale—especially in structured light and laser scanning systems \cite{salvi_pattern_2004}. As they emit their own energy, they function reliably under poor lighting, including darkness \cite{hansard_time--flight_2013}. Moreover, active sensors perform well on textureless or uniform surfaces, where passive systems typically fail \cite{curless_volumetric_1996}. However, active systems have drawbacks. Multiple active sensors can interfere when used simultaneously, reducing measurement accuracy \cite{berger_markerless_2011,zhang_microsoft_2012}. Their energy requirements for illumination limit use on mobile or low-power platforms \cite{foix_lock-time--flight_2011}. Additionally, their mechanical complexity and projection hardware contribute to higher costs compared to passive alternatives \cite{shuai_comparison_2024}.

\subsection{Passive Sensing Systems}
Passive sensing systems do not require energy projection, reducing environmental constraints and simplifying image processing \cite{ding_research_2018}. Camera-based vision systems are predominantly used in manufacturing, where they offer high-speed data acquisition with minimal infrastructure. However, their accuracy is often lower than that of active sensors and may require supplementary processing for improved precision. Common approaches include monocular and binocular vision. In general, passive sensing systems require at least two images or cameras capturing a scene from different viewpoints. The 3D structure of the scene can then be recovered through triangulation \cite{hartley_multiple_2004}.\newline

\textbf{Monocular Vision} estimates depth information from a single image by leveraging surface texture, contours, and shadow analysis \cite{scharstein_taxonomy_2001} from camera in Fig. \ref{fig:sensor_overview}. Although this method is computationally efficient, it is inherently limited by the available visual cues and lacks precise depth estimation.\newline

\textbf{Binocular Vision (Stereo Vision)} employs two cameras (\textit{cam1} and \textit{cam2} in Fig. \ref{fig:sensor_overview}) to capture images from different viewpoints, enabling depth computation through stereo matching \cite{brown_advances_2003}. This method is less affected by lighting conditions and surface textures but is sensitive to occlusions and feature matching inconsistencies.\newline

Passive sensors, such as monocular or stereo cameras, are low-cost, compact, and energy-efficient, making them ideal for embedded or mobile platforms \cite{scaramuzza_visual_2011}. They also capture rich visual data—color and texture—supporting object recognition, semantic mapping, and visual SLAM alongside 3D reconstruction \cite{szeliski_computer_2011}. As they do not emit signals, they are immune to interference, even in multi-sensor environments. Passive systems depend on ambient light, making them sensitive to low-light, glare, or shadow conditions \cite{geiger_are_2012}. Their depth estimation relies on visual correspondence, which struggles in low-texture or repetitive scenes and fails under occlusion \cite{hartley_multiple_2004}. Furthermore, monocular systems lack metric scale unless external references or assumptions are introduced \cite{fleet_lsd-slam_2014}.\newline

\textbf{Multi-Modal Approaches} - Beyond traditional contact and non-contact methods, hybrid sensing techniques and dataset-driven approaches have been explored. Multi-sensor fusion techniques combine contact-based and non-contact modalities or integrate multiple non-contact methods to improve reconstruction robustness \cite{wang_combined_2022,yu_rapid_2019,theodoropoulos_cyber-physical_2023,zhou_aviation_2024,chen_implicit_2023,wang_density-invariant_2021,liu_high-accuracy_2022}. include combining ToF with structured light or integrating laser scanning with photogrammetry. Multi-modal sensing improves 3D reconstruction by fusing complementary modalities (e.g., ToF and structured light, RGB and LiDAR), achieving high-resolution depth and robustness across surfaces and lighting conditions. Yu et al.\cite{yu_rapid_2019} present a hybrid approach integrating CMM (Contact based) with laser scanners (Non-Contact based), combining the speed of non-contact laser data acquisition with the precision of tactile probing. Laser scanners efficiently capture dense surface measurements but struggle with reflective or intricate geometries, whereas CMMs, though slower, provide superior accuracy. By leveraging laser data for broad coverage and refining it with selective probe inputs, this fusion enhances both accuracy and efficiency, supporting high-quality reverse engineering of complex industrial geometries. Nonetheless, such integration introduces challenges in system design, synchronization, and computational demand.\newline


\textbf{CAD and Dataset-Based Methods} utilize pre-existing computer-aided design (CAD) models or scanned object datasets to generate 3D reconstructions \cite{xu_fast_2017}. CAD-based approaches offer high precision by leveraging digital product representations\cite{harabin_deep_2023,li_rearrangement_2023,mu_structural_2024,jaiswal_mesh-driven_2024,liu_advancing_2024}, while dataset-based techniques enhance model robustness by incorporating real-world variations \cite{qie_function-oriented_2021}. Their limitation lies in dependency on available CAD models and weak generalization to novel or deformable objects\cite{birdal_cad_2017}. These methods are widely used in industrial inspection and digital twin applications.\newline

In conclusion, 3D reconstruction techniques continue to evolve, integrating contact-based, non-contact, and hybrid methodologies to enhance accuracy, efficiency, and adaptability. While contact-based methods provide high precision, their limitations in scalability and efficiency have led to the widespread adoption of non-contact techniques such as structured light, ToF, and laser scanning. Additionally, multi-modal approaches and CAD/dataset-driven methods further expand the capabilities of 3D reconstruction in industrial applications. As advancements in sensor technologies and computational algorithms progress, the potential for more precise, real-time, and automated 3D reconstruction solutions continues to grow, paving the way for innovative applications across various domains.

\subsection{Scanning strategy}
\label{sec:scanning_strategies}
\begin{figure}[!t]
    \centering
    \begin{subfigure}{1.0\linewidth}
        \centering
        \includegraphics[width=\linewidth]{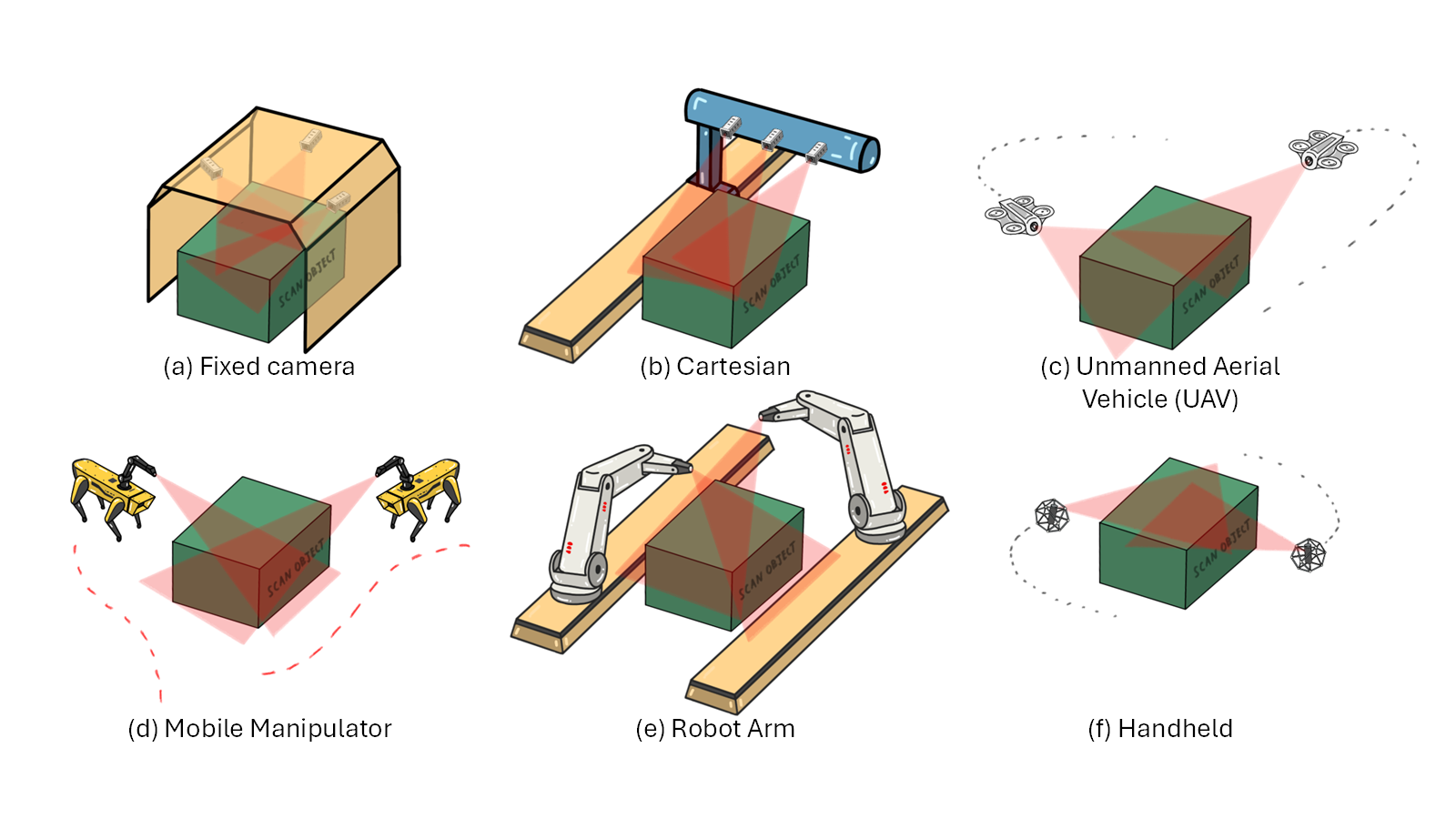}
        \caption{Automated scanning strategies: Fixed Camera, Cartesian, UAV, Robot Arm, Mobile Manipulator, Handheld.}
        \label{fig:scanningstrategy}
    \end{subfigure}
    \hfill
    \begin{subfigure}{1.0\linewidth}
        \centering
        \includegraphics[width=\linewidth]{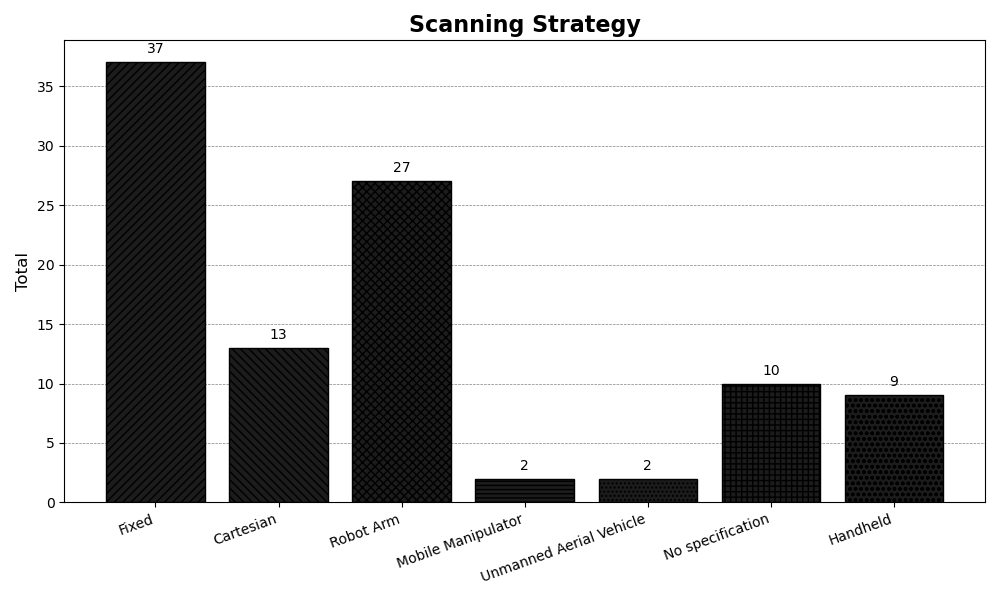}
        \caption{Overview and classification of automated scanning strategies used in 3D reconstruction.}
        \label{fig:six_type_of_scanning_strategy}
    \end{subfigure}
    \caption{Comparison of automated scanning strategies for 3D reconstruction.}
    \label{fig:scanning_strategy_comparison}
\end{figure}

Many survey papers has discussed the sensor types and it applications. However, there is no survey papers focus on discussing the scanning strategy. From the observation, scanning strategy depends on accuracy, scanning range, field of view, scan time and infrastructure cost. from the 106 papers surveyed, The 3D scanning strategy can classified into seven types (Fig. \ref{fig:scanningstrategy}): fixed camera, cartesian, robot arm, mobile manipulation, Unmanned Aerial Vehicle (UAV), handheld and hybrid system (e.g. combination of robot and fixed camera/cartesian).\newline


\textbf{Fixed camera} systems are the most popular strategy applied in review papers, which has 37\%  of the papers used it, fixed camera function as static scanning setups, employ one or more stationary cameras \cite{zhang_screening_2023,zhang_3d_2019} to capture images of static objects. Despite their utility, these systems may fail to accurately reconstruct fine details—particularly sharp edges—with optimal precision. Fixed-camera configurations can be categorized into three types: static local scanning \cite{wang_robot_2024,kumar_surface_2018,wang_digital-twin_2024,chen_deep_2024}, static local scanning with multiple viewpoints \cite{zhang_neural_2023,paulo_lima_markerless_2017,lei_boundary_2025}, and static reference marker-based scanning \cite{wang_combined_2022}. Marker-based scanning is a technique where markers are attached to the object surface and measured by devices like laser trackers and photogrammetry systems to obtain accurate 3-D coordinates as reference points for point cloud registration and pose measurement in complex assembly environments. Fixed camera provide high precision when the object is within their optimal measurement range. Nevertheless, these configurations face inherent challenges, Fixed sensors cannot easily scan complex geometries with hidden areas unless repositioned, example like blind spots in structured light scanning \cite{malhan_planning_2022} , as well as the practical difficulties associated with scanning large objects \cite{borghese_autoscan_1998}.\newline

\textbf{Cartesian systems} are the third highest strategy among 7 categories (12\%). The Cartesian system refer to gantry-based setups that allow movement along one or more axes \cite{tantivanichakit_miniature_2012}. These systems can operate using single-axis rotation or translation, or through multi-axis movements \cite{zhao_three-dimensional_2021,ding_three-dimensional_2020}. Analogous to CNC machines, a Cartesian framework moves an end effector equipped with cameras along three or four axes (translations in X, Y, Z, and/or rotation about Z), thereby enhancing scanning accuracy through closer proximity to the object. Although fixed camera and Cartesian systems are advantageous for capturing large objects in static environments, they remain limited in scanning the interiors of complex structures, such as vehicle body interiors. Moreover, despite their extended scanning range, gantry systems are prone to occlusion-related issues \cite{zhao_three-dimensional_2021}. Cartesian Coordinate Measuring Machines (CMMs) are typically designed with 3-, 4-, or 5-axis configurations, wherein each axis moves from a common zero reference. These machines are generally classified into four types: Bridge-type, Cantilever-type, Horizontal arm type, and Gantry type \cite{puertas_precision_2013,kamrani_rapid_2006,kamrani_reverse_2006}. WAAM Cartesian systems—where motions occur along orthogonal X-Y-Z axes—are straightforward to control and calibrate. They facilitate easier integration with vision sensors like laser profilometers, aligning well with machine coordinates for data acquisition \cite{huang_rapid_2022}.
WAAM systems operating purely in the X–Y plane with Z-axis growth struggle to reach complex geometries like large overhangs or inclined surfaces. As noted in tool-path planning literature \cite{huang_rapid_2022}. \newline

\begin{figure}[!t]
    \centering
        \includegraphics[width=\linewidth]{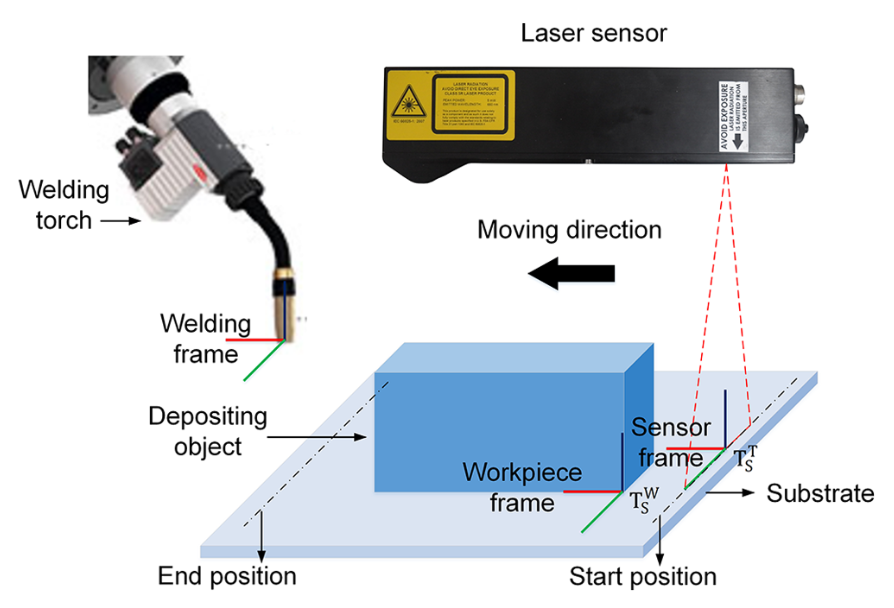}
        \caption{Coordinate transformation among the sensor, welding, and workpiece coordinate frames \cite{huang_rapid_2022}.}
        \label{fig:rapid_surface_cite}
\end{figure}

\textbf{Robot arm}-based scanning systems are the second highest strategy from the survey, which consists of 28\% of the population. Robot arm system \cite{sentenac_automated_2018,jin_automatic_2016,zhang_screening_2023} effectively address challenges in capturing curved surfaces by maintaining a stable scanning distance. Such systems typically operate with a moving local scanner \cite{nguyen_revolutionizing_2024,malhan_planning_2022,bauer_registration_2021,ren_online_2024} or by integrating the scanner onto a stationary robotic base or a mobile platform following a predetermined track \cite{zhou_aviation_2024,le_reun_improving_2024}. The benefits of robot arm-based scanning system is can reposition the sensor to cover blind spots on complex parts \cite{malhan_planning_2022}. However, these robotic systems exhibit drawbacks such as slower scanning speeds and minor vibrations, which may lead to motion blur in the captured images. In addition, singularity issues \cite{zhang_cad-aided_2022} constrain the robot’s operational range, thereby limiting its effectiveness in complex scanning tasks.\newline

\textbf{Mobile manipulators} \cite{zhou_learning-based_2022,liu_high-accuracy_2022} integrate the functionalities of fixed cameras and robot arms by combining a mobile platform with a robotic arm. This design enables the mobile manipulator carry camera or sensor to reposition data acquisition on the workpiece, thereby providing a larger field of view and enhanced flexibility in dynamic environments \cite{liu_high-accuracy_2022,zhou_learning-based_2022}. Nonetheless, structural restrictions and the need to maintain stability can limit the overall application range of this method \cite{larimi_stabilized_2013}.\newline

\textbf{Unmanned Aerial Vehicles (UAVs)} have emerged as a viable platform for 3D reconstruction, particularly in large-scale applications, as demonstrated in prior studies \cite{ding_research_2018, m_minos-stensrud_towards_2018}. By equipping UAVs with cameras or depth sensors, these systems enable the acquisition of spatial data from elevated and flexible vantage points, offering a practical solution for capturing extensive or inaccessible structures. This approach significantly improves efficiency in scenarios where conventional measurement techniques are constrained by time or physical reach \cite{ding_research_2018}. Despite these benefits, Occlusions, limited manoeuvrability in confined environments, and viewpoint planning limitations can hinder comprehensive data acquisition, thereby affecting reconstruction fidelity \cite{cao_hierarchical_2020}.\newline

\textbf{Handheld scanning} \cite{ni_three-dimension_2018,yan_point_2021,xie_aircraft_2020,fu_3d_2023,wang_trajectory_2022,peng_tool_2024} is a versatile technique that employs manually operated portable devices, offering high flexibility and an unrestricted scanning range. This method accounts for approximately 10\% of research using this strategy. Despite these advantages, it is inherently limited by the operator’s reach and physical constraints. Chew et al. \cite{chew_-process_2024} demonstrated the application of a handheld laser scanner to capture objects as ground truth; however, reconstruction time increases exponentially when high accuracy is required across numerous parts \cite{zhang_screening_2023}. Repositioning the depth sensor and integrating data from multiple viewpoints can help mitigate field-of-view limitations \cite{izadi_kinectfusion_2011}.\newline

\begin{figure}[!t]
    \centering
        \includegraphics[width=\linewidth]{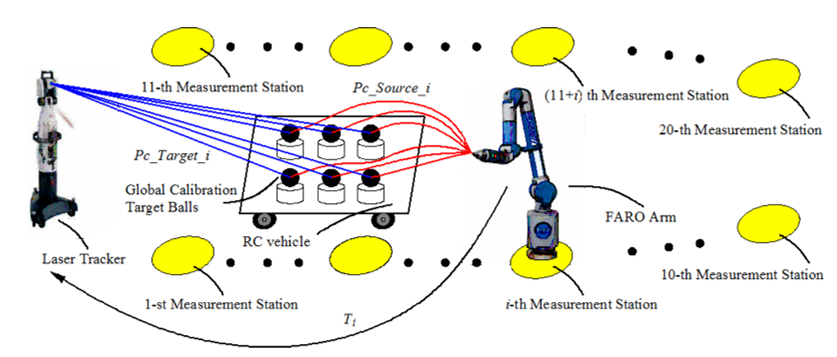}
        \caption{Schematic diagram of global calibration and station calibration \cite{zhou_aviation_2024}.}
        \label{fig:aviation_equipment_cite}
\end{figure}

\textbf{Hybrid systems} combine multiple scanning strategies to leverage their respective advantages. For example, Liu et al. \cite{liu_high-accuracy_2022} introduced a three-degree-of-freedom measurement platform using a 3D structured light camera, which provides a large field of view and high accuracy for pose measurement in large-scale robotic assembly. Their work includes a novel mechanical tooling and calibration method that utilizes eight uniformly distributed cubes and planar markers to enable efficient and precise localization of the robot’s end effector. Moreover, they propose a camera-point cloud collaborative localization technique that addresses sub-pixel marker center detection challenges to improve localization accuracy. Nonetheless, the eight-cube configuration restricts the motion space of the robot arm’s fifth joint, potentially leading to collisions during extensive rotations. Similarly, Zhou et al. \cite{zhou_aviation_2024} presented a global calibration approach employing the Faro Arm for precision measurement and station calibration to enhance reconstruction accuracy (Fig. \ref{fig:aviation_equipment_cite}), although this system constrains the robot arm's joint movement and delivers optimal performance only within a well-calibrated zone.\newline

In conclusion, various scanning methodologies for 3D reconstruction demonstrate distinct strengths and limitations. Fixed camera, Cartesian, and robotic arms each offer high precision but face challenges such as occlusion, limited operational range, and motion-induced blur. Mobile manipulators and UAVs enhance flexibility and coverage but require rigorous calibration and structural stability. Handheld and hybrid approaches provide versatility and refined localization, although at the expense of increased processing time and mechanical constraints. Future research should integrate these techniques to overcome current limitations and improve reconstruction accuracy and overall system performance.\newline

\section{Traditional 3D Reconstruction Techniques}
\label{sec:traditional_technique}

3D reconstruction is fundamental in manufacturing, supporting accurate modeling, inspection, and the development of digital twins. Among its components, {\textbf{registration is central}}, particularly in multi-view and multi-sensor settings where raw data are inherently fragmented. Without robust registration, individual captures cannot be fused into a coherent whole, and reconstruction quality is severely compromised.  

Registration provides the mathematical framework to align heterogeneous measurements within a common coordinate system, thereby ensuring metric accuracy and structural coherence. This step is indispensable in precision-driven manufacturing environments, where even sub-millimeter misalignments can lead to false defect detection or unreliable CAD comparisons.

Thus, registration functions as the backbone of the reconstruction pipeline, bridging raw sensor data with refined outputs such as surface meshes or CAD models. Its role extends across all stages, from coarse global alignment to fine local optimization. In this review, traditional 3D reconstruction techniques for manufacturing are organized into three main stages: \textit{pre-processing}, \textit{processing (coarse-to-fine registration)}, and \textit{post-processing}.

\subsection{Pre-processing}
\label{sec:pre-processing}

Pre-processing initiates the 3D reconstruction pipeline by transforming raw sensor data into geometrically structured information, laying the groundwork for accurate modeling. In manufacturing, this stage includes sensor calibration, image preprocessing, geometric feature generation, and initial structure estimation.

\textbf{Sensor Calibration and Configuration.} Accurate calibration is essential for metric reconstruction. Intrinsic calibration determines focal length, optical center, and distortion coefficients, while extrinsic calibration defines the sensor’s position/orientation relative to other devices or a world frame. Common approaches include Zhang’s chessboard-based method \cite{riffo_active_2022,li_novel_2023,xu_fast_2017}, fringe-based subpixel calibration \cite{xu_fast_2017}, and hand–eye calibration using least-squares optimization \cite{yu_rapid_2019,zhao_three-dimensional_2021,zeng_registration_2021,yang_investigation_2021,le_reun_improving_2024}. More advanced techniques such as Tsai’s dual-space algorithm \cite{zhou_learning-based_2022,sileo_vision-enhanced_2024} and the 246 algorithm \cite{zhang_robot_2019} are often applied in robotic systems for precision assembly.

\textbf{Image Rectification and Enhancement.} Preprocessing also ensures spatial consistency for stereo and multi-view setups. Rectification techniques (e.g., epipolar alignment) improve stereo correspondence \cite{jin_automatic_2016,sentenac_automated_2018}, while enhancement methods such as CLAHE \cite{Zuiderveld1994CLAHE}, binary thresholding, grayscale conversion, and RGB-to-YCrCb transformation improve feature detectability under challenging lighting conditions \cite{huang_rapid_2022,choi_vision-guided_2025,chang_eye--hand_2016}.

\textbf{Feature Detection and Description.} Robust feature extraction underpins registration and reconstruction. Detectors such as SIFT\cite{Lowe2004SIFT}, SURF\cite{Bay2008SURF}, ORB\cite{Rublee2011ORB}, and AKAZE\cite{Alcantarilla2013AKAZE} identify scale- and rotation-invariant keypoints \cite{zhang_screening_2023,zhang_cad-aided_2022,ding_research_2018,paulo_lima_markerless_2017,zheng_semantic_2025}. Shape-based techniques (e.g., Laplacian filtering, edge/circle detection \cite{mu_structural_2024,huang_database-assisted_2021,xu_3d_2022,huang_rapid_2022,choi_vision-guided_2025}) and descriptors such as FPFH \cite{nigro_assembly_2023} or Harris corners \cite{shalaby_mathematically_2022} enable robust geometric matching.

\textbf{Feature Matching and Correspondence.} Correspondences between views are established by comparing descriptors. RANSAC \cite{fischler_random_1981} remains the standard for outlier rejection \cite{nigro_assembly_2023,zhang_screening_2023,shalaby_mathematically_2022,wang_digital-twin_2024,wang_density-invariant_2021,liu_approach_2024}, while FLANN \cite{zhang_cad-aided_2022,reno_comparative_2020}, HRNet \cite{jiang_geometry_2024}, and ISS \cite{kinnell_autonomous_2017} provide fast or specialized matching.

\begin{figure}[!t]
    \centering
    \includegraphics[width=\linewidth]{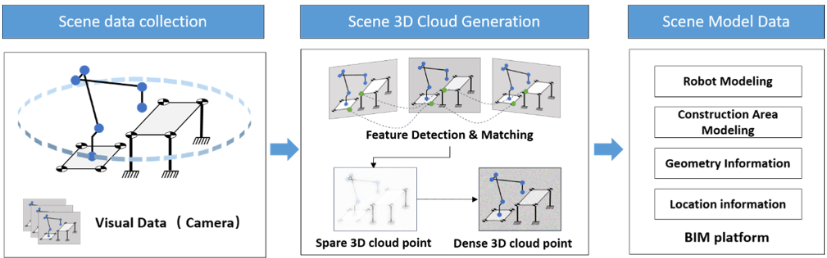}
    \caption{Sparse and dense point cloud generation for the robot brick BIM task model \cite{ding_bim-based_2020}.}
    \label{fig:robot_brick_cite}
\end{figure}

\textbf{Pose Estimation and Structure Generation.} Using intrinsic/extrinsic parameters and feature correspondences, relative poses are estimated via the essential or fundamental matrix. This enables triangulation for sparse 3D reconstruction through Structure from Motion (SfM) \cite{zhang_screening_2023,yang_dpps_2023,ding_bim-based_2020,petruccioli_assessment_2022} or CMVS \cite{ding_bim-based_2020}. Dense reconstructions are generated via MVS algorithms such as COLMAP \cite{schonberger_structure--motion_2016,wang_robot_2024,hu_point_2024,jaiswal_mesh-driven_2024} or PMVS \cite{furukawa_accurate_2010,zhang_cad-aided_2022,zhang_screening_2023,ding_research_2018}. For example, Ding et al. \cite{ding_bim-based_2020} reconstructed a robotic brick-picking scene using a pipeline of calibration, feature matching, SfM, and MVS to produce a dense point cloud for BIM integration (Fig.~\ref{fig:robot_brick_cite}).

\subsection{Processing: Coarse-to-Fine Registration}
\label{sec:registration}

At the core of multi-view and multi-sensor 3D \newline reconstruction lies \textbf{registration}—the process of aligning datasets into a unified coordinate frame. In manufacturing, achieving accurate registration is critical: even sub-millimeter misalignments can propagate into dimensional errors that fail quality checks.

A widely adopted strategy is \textbf{coarse-to-fine registration}, executed in two stages:

\begin{itemize}
    \item \textbf{Coarse (Global) Registration} estimates an initial transformation between point clouds without prior pose information. It must handle large misalignments and unknown viewpoints.
    \item \textbf{Fine (Local) Registration} refines this transformation, reducing residual errors to meet manufacturing tolerances.
\end{itemize}

\noindent\textbf{Why coarse?}  
Coarse registration provides robustness against large initial pose differences. In manufacturing, components may be scanned from opposite sides or with significant displacement; direct fine alignment would fail without a reasonable initial guess. Methods include RANSAC-based feature matching \cite{zhang_screening_2023,mosca_ransac-based_2020}, FPFH-based correspondence search \cite{nigro_assembly_2023}, and pose graph optimization \cite{jin_automatic_2016}. For example, Yu et al. \cite{yu_rapid_2019} applied coarse registration to align large-part scans before robotic assembly.

\noindent\textbf{Why fine?}  
Fine registration achieves the precision required for inspection or reverse engineering. Starting from the coarse estimate, algorithms such as ICP \cite{besl_method_1992,yan_point_2021}, point-to-plane ICP \cite{nigro_assembly_2023,wang_digital-twin_2024}, or customized ICP \cite{iuso_cad-based_2021,fu_3d_2023} iteratively minimize geometric discrepancies. Nigro et al. \cite{nigro_assembly_2023} demonstrated that a coarse FPFH alignment followed by point-to-plane ICP yielded the sub-millimeter accuracy needed for peg-hole verification.

\begin{table*}[t]
\centering
\caption{Comparison between global (coarse) and local (fine) registration methods in 3D reconstruction}
\label{tab:coarse_fine_registration}
\resizebox{\textwidth}{!}{
\begin{tabular}{p{3.5cm}p{6.5cm}p{6.5cm}}
\hline
\textbf{Aspect} & \textbf{Global Alignment (Coarse Registration)} & \textbf{Local Alignment (Fine Registration)} \\ \hline\hline

\textbf{Objective} 
& Establish an initial transformation between unaligned point clouds or views 
& Refine the transformation to minimize alignment error between corresponding points \\ 

\textbf{Principle} 
& Uses geometric or descriptor-based correspondences to compute transformation hypotheses 
& Minimizes point-to-point or point-to-plane distances iteratively \\ 

\textbf{Typical Techniques} 
& RANSAC-based matching \cite{zhang_screening_2023,shalaby_mathematically_2022,mosca_ransac-based_2020,yu_rapid_2019,kinnell_autonomous_2017,sileo_vision-enhanced_2024,peng_tool_2024}, FPFH feature matching \cite{nigro_assembly_2023,wang_digital-twin_2024}, pose graph optimization \cite{jin_automatic_2016,zhao_position_2019} 
& ICP \cite{yan_point_2021,theodoropoulos_cyber-physical_2023,hu_point_2024,chew_-process_2024,xu_fast_2017,chang_vision-based_2021,zeng_registration_2021}, point-to-plane ICP \cite{nigro_assembly_2023,wang_digital-twin_2024}, customized ICP \cite{iuso_cad-based_2021,fu_3d_2023,skeik_6d_2022}, SVD \cite{xu_3d_2022,zhao_position_2019} \\ 

\textbf{Data Requirements} 
& Requires distinctive geometric features; no initial pose needed 
& Needs a good initial guess from global registration to converge \\ 

\textbf{Strengths} 
& Robust to large misalignments; enables automatic alignment from arbitrary views 
& High accuracy in structured or smooth regions \\ 

\textbf{Limitations/Challenges} 
& Sensitive to outliers or repetitive patterns 
& Prone to local minima without good initialization \\ 

\textbf{Application Scenarios} 
& First-pass alignment, integration of disconnected scans 
& Final alignment, dimensional inspection, reverse engineering \\ 

\hline
\end{tabular}
}

\end{table*}

\subsection{Post-processing}
\label{sec:post-processing}

Post-processing refines the registered dense point cloud into a usable 3D model through cleaning, meshing, and formatting.

\textbf{Noise Reduction and Outlier Removal.} Statistical filtering \cite{wang_application_2022,fu_3d_2023,liu_approach_2024,huang_rapid_2022,peng_tool_2024}, radius-based thresholding \cite{yang_dpps_2023,dai_graspnerf_2023,zhang_robot_2019,song_two-stage_2021}, and RANSAC filtering \cite{liu_improved_2019,xu_3d_2022,mosca_ransac-based_2020,wang_pose_2020} improve point cloud quality. Bilateral filtering can smooth noise while preserving edges \cite{ito_free-viewpoint_2024,wang_application_2022}.

\textbf{Resampling and Downsampling.} Voxel grid filtering reduces point density while maintaining shape \cite{fu_3d_2023,liu_approach_2024,m_minos-stensrud_towards_2018}. Adaptive sampling strategies balance computational load and detail retention \cite{li_c-lfnet_2024}.

\textbf{Surface Reconstruction and Meshing.} Poisson \newline reconstruction \cite{wang_robot_2024,ding_research_2018}, Delaunay triangulation \cite{nigro_peg--hole_2020,ito_free-viewpoint_2024}, and TSDF fusion \cite{dai_graspnerf_2023,chew_-process_2024} produce watertight meshes. Commercial tools such as Geomagic and Metashape \cite{petruccioli_assessment_2022,ni_three-dimension_2018} are common in industry.

\textbf{Geometric Segmentation.} Segmentation isolates functional parts via primitive fitting (RANSAC) \cite{wang_pose_2020,yu_rapid_2019} or CNN-based classification \cite{qie_function-oriented_2021,skeik_6d_2022}.

\textbf{Representation and Output Formatting.} Outputs may be explicit (meshes) for CAD/CAM, implicit (SDF, occupancy grids) for learning-based systems, or hybrid for interoperability.

\section{Machine Learning/Deep Learning Reconstruction Approaches}
\label{sec:ml_dl_approaches}

The integration of machine learning (ML) and deep learning (DL) techniques into 3D reconstruction pipelines has revolutionized quality assurance and process optimization in smart manufacturing systems. Unlike traditional photogrammetric methods requiring precise camera calibration and controlled lighting, data-driven approaches learn implicit relationships between 2D sensor data and 3D geometries through exposure to diverse training samples. This paradigm shift enables robust reconstruction of complex industrial components from single-view imagery - a critical capability for inline inspection systems with space constraints.Some of performance comparisons of DL-based 3D reconstruction methods in manufacturing applications are shown in Table \ref{tab:verified_methods}.

\subsection{Point Cloud Processing with Deep Networks}
Recent advances in point cloud processing have enabled more robust 3D reconstruction in dynamic manufacturing environments. Zhou et al. \cite{zhou2022learning} demonstrated the effectiveness of PV-RCNN for industrial object detection, combining voxel-based sparse convolutions with point-based feature aggregation to handle objects spanning three orders of magnitude in size (from 5mm screws to 9.6m workstations). Their approach achieved 82\% success in 6-DoF peg-in-hole tasks, showcasing the potential of hybrid architectures for manufacturing applications. Chen et al. \cite{chen_implicit_2023} further advanced this domain through neural signed distance functions (SDFs) that fuse multi-sensor data, reducing reconstruction errors by 69\% in complex S-shaped test parts compared to single-sensor baselines.

\subsection{Generative Architectures for Shape Inference} 

Generative adversarial networks (GANs) \cite{goodfellow2014generative} demonstrate considerable potential in manufacturing by synthesizing plausible height maps from sparse training data. In particular, Wasserstein GAN formulations with gradient penalty (WGAN‑GP) \cite{gulrajani2017improved} mitigate mode collapse issues—prevalent in early GAN variants—by enforcing a 1‑Lipschitz constraint on the critic via a gradient‑norm penalty on interpolated samples between real and generated distributions. This smoothes the optimization landscape, providing more informative gradients and stabilizing training, which leads to improved diversity and convergence stability even when reconstructing intricate features of laser‑welded hairpin connectors, achieving structural similarity indices (SSIM) of 0.87 on validation sets.

Beyond adversarial frameworks, Wang \textit{et al.} \cite{wang2020shape} introduced a point cloud completion method that leverages shape priors through a variational autoencoder (VAE) combined with graph neural networks (GNNs). Their coarse-to-fine pipeline aligns latent features of partial inputs with those of complete shapes using feature matching and adversarial losses, followed by a refinement stage optimized with Chamfer Distance. This approach enables accurate recovery of occluded or missing regions in 3D scans, addressing challenges in manufacturing contexts where sensor limitations and restricted viewpoints frequently degrade reconstruction quality.  

The adversarial framework proves even more effective when integrated with neural radiance fields (NeRF). For instance, Liu et al.\ \cite{liu2023rgbgrasp} introduced the RGBGrasp framework, which integrates hash encoding with monocular depth priors and reports an 82.2\% grasp success rate in cluttered environments—surpassing conventional RGB‑D fusion methods by 34\%. This underscores the strength of hybrid generative–hybrid representations for advanced manufacturing tasks.

\subsection{U-Net Variants for Precision Manufacturing}
Generative models employing symmetric encoder–decoder structures with skip connections remain prevalent in the literature due to their sample efficiency on small industrial datasets. A notable example is the stacked dilated U‑Net (SDU‑Net) \cite{wang2020sdu}, a U‑Net variant that replaces the standard double convolution blocks with a combination of one standard convolution followed by multiple dilated convolutions, concatenated to the next operation; this architecture achieved comparable or superior segmentation performance using only around 40\% of the trainable parameters required by the original U‑Net. While SDU‑Net has been primarily evaluated on medical imagery (e.g., ultrasound and MRI scans), its efficiency gains—specifically, a marked reduction in parameters coupled with maintained accuracy—suggest promising applicability to reconstructing stamped metal components in manufacturing.

Further enhancements arise from integrating attention mechanisms. Qie et al.\ \cite{qie2021function} introduced a CNN-based model leveraging attention for inferring assembly relationships and kinematic constraints, achieving 91\% accuracy in gearbox component function labeling. This underscores how attention modules can effectively capture both geometric and functional semantics, aligning with a function-aware reconstruction paradigm.

\subsection{Autoencoder-Based Dimensional Metrology}
Stacked autoencoder (SAE) architectures \cite{vincent2010stacked} provide computationally efficient solutions for high-speed inspection lines. By disentangling geometric features in bottleneck representations, SAEs achieve 98.7\% classification accuracy on thread engagement verification tasks - outperforming manual CMM measurements in cycle time and repeatability. The coarse-to-fine learning strategy employed by Lu et al. \cite{lu_cfvs_2023} complements these approaches. Their CFVS system used separate networks for coarse pose estimation (OAKN) and fine alignment (OPN), achieving 100\% success in 3-DoF peg-in-hole tasks with 30cm initial errors. This hierarchical approach could enhance autoencoder-based systems dealing with large dimensional variations.

Wang \textit{et al.} (2020)~\cite{wang2020shape} present a deep learning-based framework for point cloud completion that leverages learned shape priors to recover missing or occluded regions in partial 3D scans, a common challenge in real-world manufacturing due to sensor occlusions and limited viewpoints. Their approach employs a coarse-to-fine pipeline, beginning with the training of an auto-encoder on complete point clouds to obtain robust latent embeddings (256- and 1024-dimensional) that capture global shape characteristics. These priors guide the completion network, which aligns features from partial inputs to full-shape representations via two complementary objectives: (i) an L\(_2\) feature matching loss to minimize latent space discrepancies, and (ii) an adversarial loss based on Maximum Mean Discrepancy (MMD-GAN) to reduce distributional gaps in a reproducing kernel Hilbert space. A refinement stage then concatenates the coarse output with the original partial input, optimizing jointly with Chamfer Distance, feature alignment, and adversarial losses to generate dense, high-fidelity reconstructions. Experimental results on synthetic and real-world datasets demonstrate superior reconstruction accuracy compared to prior methods.

\subsection{Multi-Modal Fusion Architectures}
State-of-the-art systems increasingly combine neural rendering with traditional geometric pipelines. The Atlas framework \cite{murez_atlas_2020} integrates monocular depth prediction networks with volumetric TSDF fusion, enabling automatic alignment of reconstructed CAD models to optical inspection data.
Recent innovations in implicit neural representations (INRs) have been particularly impactful. Chen et al.'s \cite{chen_implicit_2023} work on gradient-constrained implicit functions demonstrates how these representations can handle complex surfaces that defy explicit parameterization, while Qie et al. \cite{qie2021function} showed their potential for incorporating manufacturing semantics through rule-based reasoning.
\begin{table*}[htbp]
\centering
\caption{Performance of some DL-based 3D reconstruction methods in manufacturing applications}
\label{tab:verified_methods}
\resizebox{\textwidth}{!}{
\begin{tabular}{lllr}
\hline
\textbf{DL Method} & \textbf{Application} & \textbf{Reported Metric} & \textbf{Value} \\
\hline
PV-RCNN \cite{zhou2022learning} & Hybrid object detection & 6-DoF peg-in-hole success rate (6-DoF) & 82\% \\
Neural SDF fusion \cite{chen_implicit_2023} & Multi-sensor surface reconstruction & Reconstruction error reduction vs single-sensor & 69\% \\
WGAN-GP \cite{gulrajani2017improved} & Height map generation (laser-welded hairpin connectors) & Structural Similarity Index (SSIM) & 0.87 \\
RGBGrasp \cite{liu2023rgbgrasp} & Grasping in cluttered scenes & Grasp success rate improvement over baselines & +34\% \\
SDU-Net \cite{wang2020sdu} & Sample-efficient segmentation architectures & Parameter count reduction vs U-Net & ~60\% \\
CFVS \cite{lu_cfvs_2023} & Coarse-to-fine 3-DoF peg-in-hole & Success rate (3-DoF) & 100\% \\
Stacked Autoencoder (SAE) \cite{vincent2010stacked} & Thread engagement verification & Classification accuracy & 98.7\% \\
\hline
\end{tabular}
}

\end{table*}
\subsection{Multilayer Perceptrons (MLP)}
Neural Radiance Fields (NeRF) \cite{mildenhall_nerf_2020} encode scenes as continuous volumetric functions using multilayer perceptrons (MLPs). Given 3D coordinates and viewing directions, the network simultaneously predicts \textit{volume density} ($\sigma$) and \textit{view-dependent color} ($c$). The model is trained via differentiable volume rendering, where rays are sampled through scene pixels and optimized using a photometric loss between synthesized and real images. This enables high-fidelity novel view synthesis but requires extensive per-scene optimization and lacks explicit surface geometry.

\begin{figure}[!t]
    \centering
    \includegraphics[width=\linewidth]{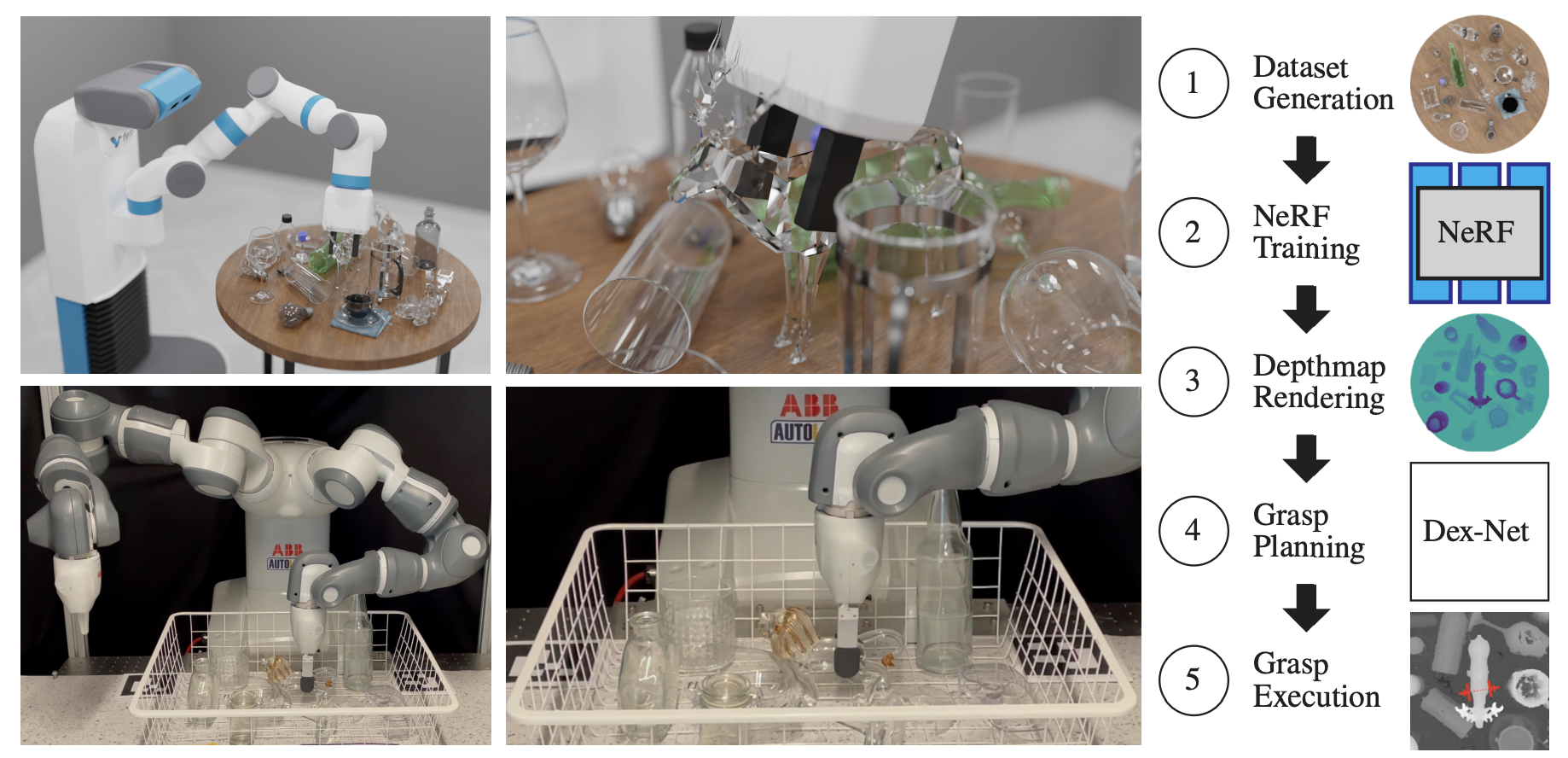}
    \caption{Using NeRF to grasp transparent objects Given a scene with transparent objects (left column), the pipeline is on the right to compute grasps (middle column). \cite{ichnowski_dex-nerf_2021}.}
    \label{fig:Dex-nerf}
\end{figure}

A notable example of extending NeRF beyond novel view synthesis is \textit{Dex-NeRF}~\cite{ichnowski_dex-nerf_2021}, which integrates NeRF into a robotic grasp-planning pipeline for transparent objects (Fig.~\ref{fig:Dex-nerf}). By leveraging the view-dependent volumetric representation of NeRF, Dex-NeRF can accurately reconstruct scene geometry even in the presence of transparency—a scenario that typically challenges conventional depth sensors. The system requires only calibrated RGB images with known camera intrinsics and extrinsics, making it particularly suitable for fixed workcells or robot arms with accurate kinematic encoders. Experimental results demonstrate that Dex-NeRF achieves over 90\% grasp success rates in both synthetic and real-world settings, highlighting the potential of NeRF-based MLP architectures in practical robotic manipulation tasks.

\subsection{Gaussian Splatting}
Gaussian Splatting \cite{dalal_gaussian_2024} represents scenes as collections of 3D Gaussians, each parameterized by position ($\mu$), covariance matrix ($\Sigma$), opacity ($\alpha$), and spherical harmonic coefficients for view-dependent color. During rendering, Gaussians are projected to 2D and sorted by depth, with pixel colors computed via alpha-blending along viewing rays (A compositing technique where colors of overlapping Gaussians are weighted by their opacity values and accumulated along the ray, effectively simulating light attenuation and transparency during rendering).
This differentiable approach achieves real-time performance by leveraging GPU rasterization, making it ideal for interactive applications, though it prioritizes visual fidelity over physical accuracy.\newline

\subsection{Comparative Analysis and Future Outlook}
Deep learning approaches fundamentally redefine 3D reconstruction paradigms in manufacturing compared to traditional geometric methods. While conventional techniques like structured-light scanning and stereo photogrammetry rely on physical principles of light projection and triangulation, data-driven methods learn implicit reconstruction priors from training corpora. This distinction manifests in three critical trade-offs: (1) Traditional methods provide mathematically traceable measurements with micron-level repeatability but require controlled environments and multi-view setups, whereas deep learning enables single-view inference in dynamic production settings at the cost of interpretability; (2) Physics-based approaches excel at reconstructing specular surfaces through controlled illumination patterns, while neural networks better handle occluded geometries and material variations through learned context completion; (3) Hardware-intensive traditional systems achieve deterministic real-time performance (ms latency), contrasting with DL's software-centric flexibility at higher computational costs (seconds-scale inference).

The emergence of hybrid architectures suggests a convergent future where neural networks initialise reconstructions for subsequent physical refinement - blending the adaptability of learned representations with the precision of geometric principles. This synthesis addresses manufacturing's dual demands for robust in-situ quality control and metrological-grade certification. Future advancements must reconcile scalability challenges, as DL's data hunger conflicts with manufacturers' proprietary constraints, while traditional methods struggle with Industry 4.0's demand for embedded intelligence. The field's ultimate trajectory points toward physics-informed neural networks that encode manufacturing semantics - moving beyond geometric reconstruction to functional intent understanding.

\section{Applications}
\label{sec:applications}
From the 106 reviewed articles, the applications of 3D reconstruction in the manufacturing industry can be categorized into two perspectives: process-specific focus and industrial-specific focus.
\subsection{Process Specific Focus}

\begin{figure}[!t]
    \centering
        \includegraphics[width=0.48\textwidth]{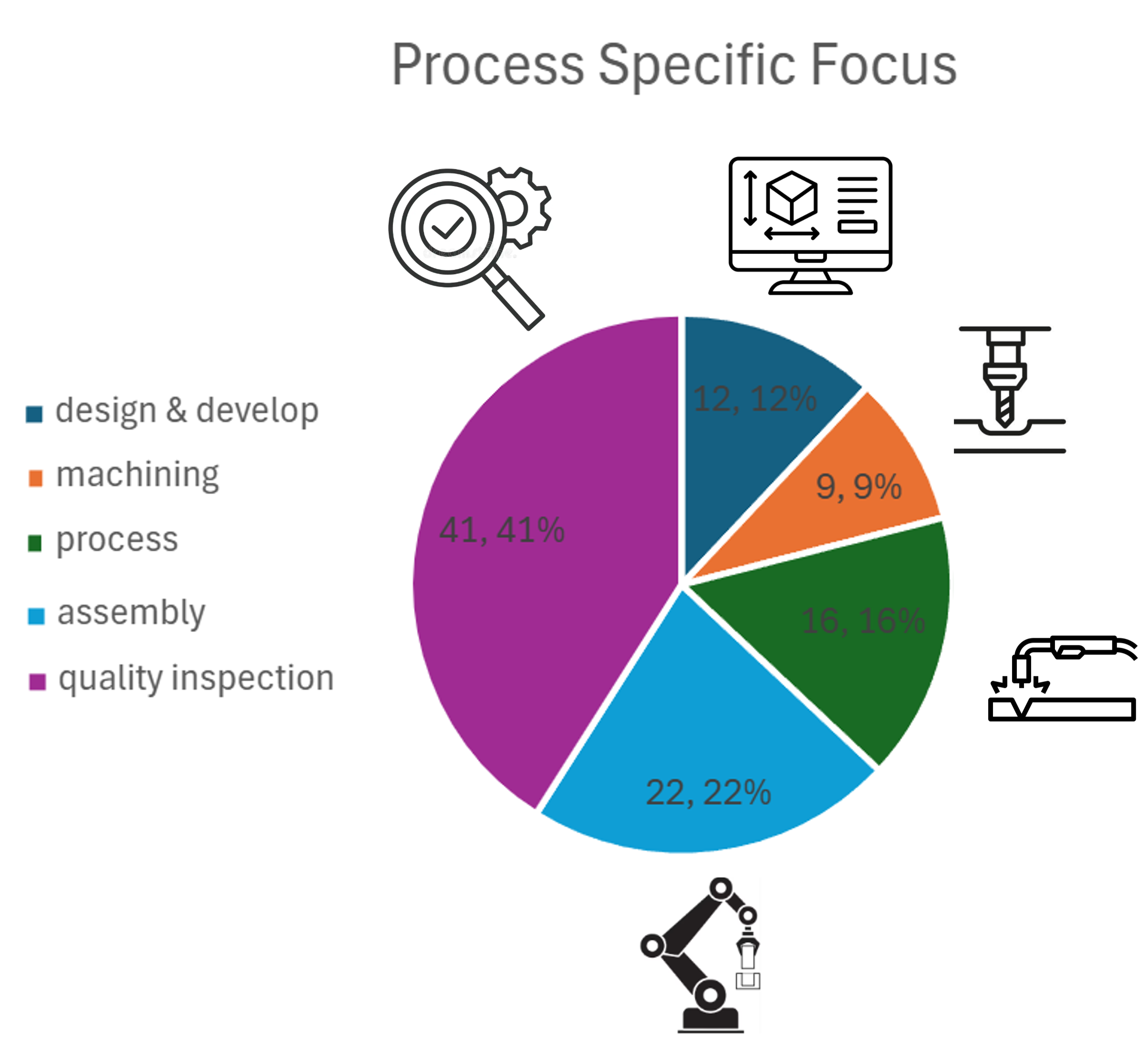}
        \caption{Process Specific Focus. Where design and development
(12), machining (8), process (17), assembly (22), and quality
inspection (41)}
        \label{fig:processspecific}
\end{figure}

 \textbf{Process-specific focus} (Fig. \ref{fig:processspecific}) perspective delineates the utilization of 3D reconstruction across manufacturing domains, encompassing design and development (12), machining (8), process (17), assembly (22), and quality verification (41). In each segment, reconstruction methodologies are leveraged to elevate operational accuracy, streamline automation, and increase throughput, with quality assessment representing the most widely implemented application. Design and development focus on improving the design and development process, including reverse engineering \cite{ni_three-dimension_2018,yan_point_2021,falheiro_cad_2021}, rapid prototyping\cite{zhang_screening_2023}, and CAD design evaluation\cite{qie_function-oriented_2021,gao_3d_2018}. Machining involves the use of 3D reconstruction techniques to enhance fabrication quality or improve process time. This category includes material processing, example like grinding \cite{zhang_robot_2019,xie_measurement_2024}, milling and lathe \cite{wang_trajectory_2022,harabin_deep_2023,ding_robust_2019} and glueing \cite{prezas_multi-purpose_2024}. The process is a systematic series of actions designed to achieve a specific objective. In manufacturing, processes like welding \cite{zhang_3d_2019,liu_approach_2024}, additive manufacturing \cite{huang_database-assisted_2021,zhou_learning-based_2022}, and stamping \cite{attar_development_2023} transform raw materials into finished products efficiently, ensuring quality and consistency. Assembly refers to the process of connecting single or multiple parts together, including part mating \cite{jiang_geometry_2024,nigro_assembly_2023,he_fast_2020} and peg-in-hole \cite{nigro_peg--hole_2020}, fastening \cite{xu_fast_2017}, pose estimation \cite{pas_grasp_2017}, palletizing\cite{ding_bim-based_2020} and installation \cite{nguyen_revolutionizing_2024}. The last category is the Quality inspection, quality inspection is the largest category amongst all 3D reconstruction applications from the survey. The advantages of 3D reconstruction applications in this area include improving or verifying product quality by evaluating scanned object properties (position, shape, size) \cite{shalaby_mathematically_2022,wang_combined_2022,zhao_position_2019}, scene alignment\cite{petschnigg_point_2020} and defects \cite{yan_point_2021,huang_rapid_2022,fu_3d_2023}against ground truth using various analysis tools. The details of the analysis tools can be found in Table \ref{tab:Analysis tool}.\newline

\begin{table}[h]
    \centering
    \caption{Analysis tool: Methods and algorithms with citation papers}
    \label{tab:Analysis tool}
    
    \resizebox{\columnwidth}{!}{
    \begin{tabular}{ll}
    \hline
    \textbf{Method and Algorithm} & \textbf{Citation}                  \\ \hline
    Polyworks\cite{noauthor_polyworks_2023} & \cite{yan_point_2021,zhou_aviation_2024,yang_investigation_2021,zhao_position_2019,zheng_laser_2025} \\ 
    CloudCompare\cite{noauthor_cloudcompare_2023} & \cite{chew_-process_2024}          \\ 
    Meshlab\cite{cignoni_meshlab_2008} & \cite{wang_robot_2024}  \\ 
    Geomagic\cite{Geomagic2024} & \cite{yu_rapid_2019,zhao_three-dimensional_2021,sentenac_automated_2018,zeng_registration_2021,ding_robust_2019,xia_accurate_2025,petruccioli_assessment_2022} \\ 
    MATLAB\cite{MATLAB2024} & \cite{ding_three-dimensional_2020,kim_highly_2024} \\  
    COMSOL\cite{COMSOL2024} & \cite{liu_advancing_2024}   \\ \hline
    \end{tabular}
    }
    
\end{table}

The analysis tools presented in Table \ref{tab:Analysis tool} highlight the prevalence of various methods for evaluating reconstructed results against ground truth. Several well-established algorithms have been integrated into commercial tools. For instance, classic approaches such as Multiview Stereo (MVS) \cite{furukawa_multi-view_2015} and Structure from Motion (SfM) \cite{schonberger_structure--motion_2016} have undergone continuous refinement to improve accuracy, efficiency, and robustness. These foundational techniques have led to the development of widely used applications, including COLMAP \cite{schonberger_structure--motion_2016}, OpenMVG \cite{kerautret_openmvg_2017}, and Visual SfM \cite{wu_visualsfm_2011}, which facilitate features to point cloud generation.

For advanced processing, tools such as MeshLab \cite{cignoni_meshlab_2008} and CloudCompare \cite{noauthor_cloudcompare_2023} provide integrated functionalities, including mesh reconstruction, 3D model generation, and scale adjustments. Additionally, industrial-grade solutions such as PolyWorks \cite{noauthor_polyworks_2023}  offer precise analytics and reporting features, catering to professional and commercial applications.\newline

\subsection{Industry-Specific Focus}

\begin{figure}[!t]
    \centering
        \includegraphics[width=0.48\textwidth]{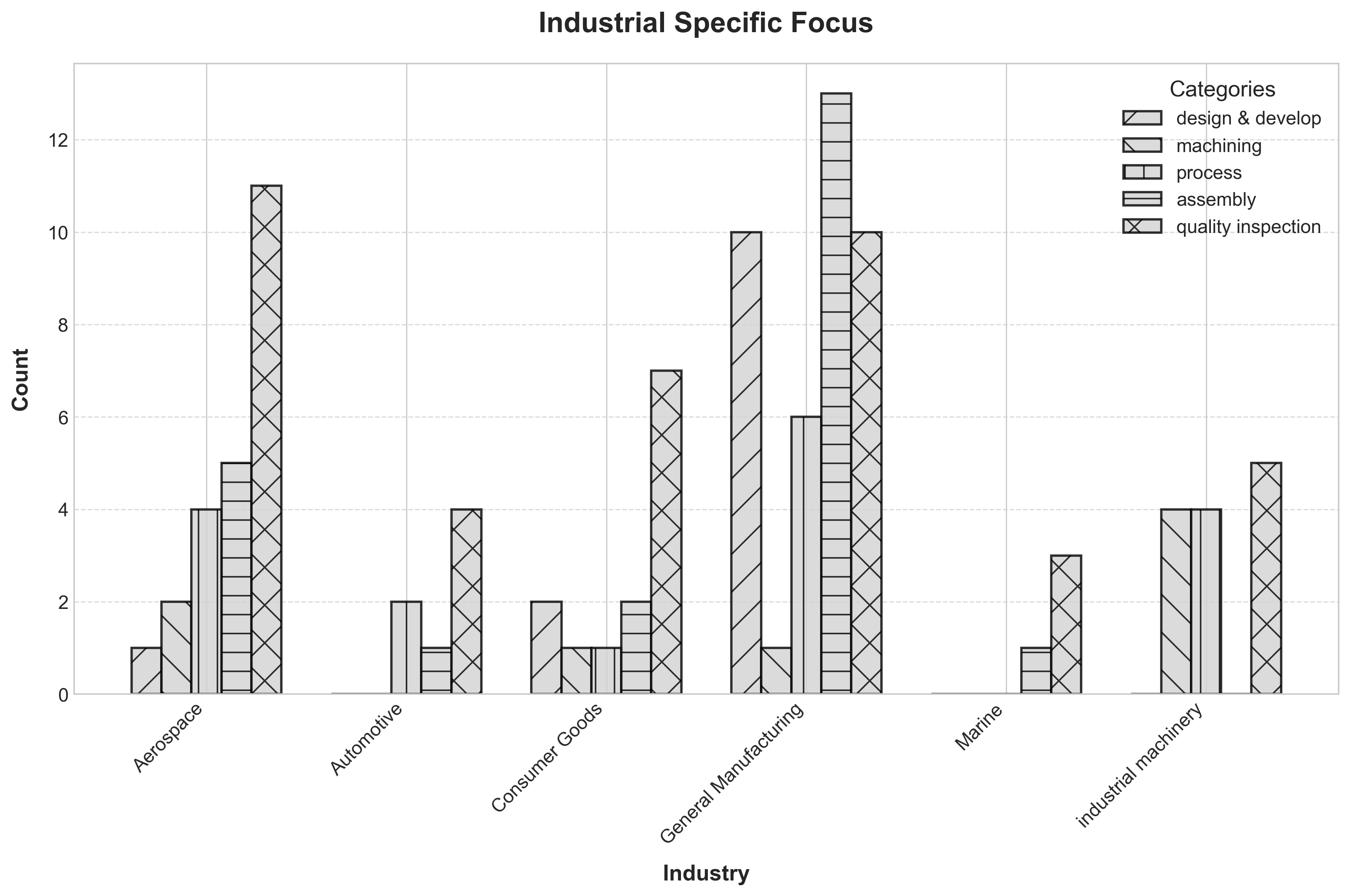}
        \caption{Industrial specific Focus. Where aerospace (23), automotive (7), consumer goods (13), general manufacturing (40), marine (4) and industrial machinery (13).}
        \label{fig:industrial specific}
\end{figure}

\textbf{Industry-specific focus} (Fig. \ref{fig:industrial specific}) highlights the deployment of 3D reconstruction technologies demonstrates significant variability across sectors such as aerospace (23), automotive (7), consumer products (13), general manufacturing (40), marine engineering (4), and industrial machinery (13). Sector-specific integration patterns reveal distinct emphasis, with general manufacturing prioritizing design refinement and inspection, automotive and aerospace industries adhering to stringent quality standards, and marine as well as machinery domains adopting reconstruction tools for specialized, targeted use-cases.\newline

Each industrial domain imposes unique requirements on 3D reconstruction systems. In semiconductor inspection, sub-micrometer vertical resolution from WLI or confocal microscopy is crucial. Automotive body measurement demands wide-area, robot-mountable structured light systems like GOM  \cite{zeiss_optical_3d_scanning}, prioritizing scanning speed and repeatability. Aerospace applications often favor CT or laser triangulation with deep penetration and high fidelity for composite and metallic parts. These variations influence sensor selection and scanning strategy based on factors such as field of view, surface reflectivity, throughput, and cost.\newline

\textbf{General Manufacturing} sector exhibits the highest activity in 3D reconstruction, covering a wide range of manufacturing processes. A key focus is on design and development (9), with Ni et al. \cite{ni_three-dimension_2018} reconstructing parts to analyze design intent or support remanufacturing. In assembly (14), 3D reconstruction aids object pose estimation, improving robotic assembly through geometry and motion planning \cite{jiang_geometry_2024} and collision avoidance \cite{harabin_deep_2023}. Quality inspection (10) is essential for defect detection \cite{riffo_active_2022}. Chang et al. \cite{chang_vision-based_2021} proposed a visual servoing approach with point cloud registration to reduce human intervention. Additionally, Zhang et al. \cite{zhang_cad-aided_2022} introduced a CAD-aided 3D reconstruction method based on time-series analysis to reduce inspection processing time.\newline

\textbf{Consumer Goods} sector focuses on process optimization to address rapid market shifts and shorter product life cycles, requiring speed and flexibility in production \cite{ni_three-dimension_2018}. Quality inspection (8) is crucial for identifying defects during assembly \cite{yan_point_2021}. Breyer et al. \cite{breyer_volumetric_2021} introduced the Volume Grasping Network (VGN), which enables real-time quality prediction and gripper positioning. Yan et al. \cite{yan_point_2021} proposed a point cloud-based reverse engineering technique for early defect detection. Fu et al. \cite{fu_3d_2023} developed an ICP alignment optimization method to enhance PCB height defect detection accuracy.\newline

\textbf{Automotive} sector places strong emphasis on quality inspection (4), driven by stringent safety and emissions regulations \cite{zhang_screening_2023}. Manufacturers implement advanced inspection protocols and testing technologies to ensure compliance. Petruccioli et al. \cite{petruccioli_assessment_2022} introduce a close-range photogrammetry-based assessment method to compare 3D acquisitions with ground truth for car body quality evaluation. Zhang et al. \cite{zhang_neural_2023} propose neural rendering-enabled 3D modeling for the rapid digitalization of in-service vehicles. Petschnigg et al. \cite{petschnigg_point_2020} leverage deep learning to automate automotive assembly simulation generation, streamlining the manufacturing process.\newline

\textbf{Aerospace} sector places significant emphasis on quality inspection (11), driven by stringent safety requirements and high-performance expectations \cite{wang_combined_2022, zhou_aviation_2024}. Wang et al. \cite{wang_density-invariant_2021} propose a global registration framework for aircraft inspection, aligning multiple point clouds with target detection and hierarchical optimization. Yang et al. \cite{yang_investigation_2021} introduce uncertainty-based point cloud registration for gap measurement in aircraft wing assembly.

For robot machining, Wang et al. \cite{wang_trajectory_2022} optimize trajectory planning based on measured point clouds, addressing large-part deformation. Tan et al. \cite{tan_multi-robot_2023} develop a 3D perception system to tackle measurement and assembly challenges across different workstations. Le Reun et al. \cite{le_reun_improving_2024} introduce robotics-based X-ray Computed Tomography for large-scale internal part reconstruction.

For assembly (5), precise 3D pose estimation is essential for multi-robot assembly \cite{wang_pose_2020}. Xie et al. \cite{xie_aircraft_2020} focus on rivet fitting detection for quality assessment, while Wang et al. \cite{wang_pose_2020} apply RANSAC cylinder fitting to extract pose estimations for large cabin assembly. These studies highlight the growing role of 3D reconstruction in improving quality inspection, machining, and assembly processes in aerospace manufacturing.\newline

\textbf{Industrial Machinery} sector emphasizes all three 3D reconstruction application categories. Since equipment is often customized to client specifications, visual inspection is crucial for ensuring quality \cite{huang_rapid_2022}. Additionally, machining relies on accurate surface information for robot path planning, as robots lack the necessary flexibility and adaptability \cite{zhang_3d_2019}. Huang et al. \cite{huang_rapid_2022} introduced quality monitoring for wire arc additive manufacturing (WAAM) using a laser profiler and support vector machine (SVM) model, demonstrating 3D reconstruction’s role in enhancing precision and process control.\newline

\textbf{Marine} sector has the lowest adoption of 3D reconstruction applications, primarily focusing on quality inspection to address maritime challenges. Shipbuilding and marine equipment manufacturing require specialized training and are time-intensive \cite{ding_research_2018}. To improve efficiency, Wang et al. \cite{wang_application_2022} applied 3D laser scanning in hull rib assembly, enabling rapid error detection and correction. Kim et al. \cite{kim_highly_2024} introduced laser trackers for ship block assembly, replacing labor-intensive traditional measurement methods.\newline

\begin{figure}[!t]
    \centering
        \includegraphics[width=0.48\textwidth]{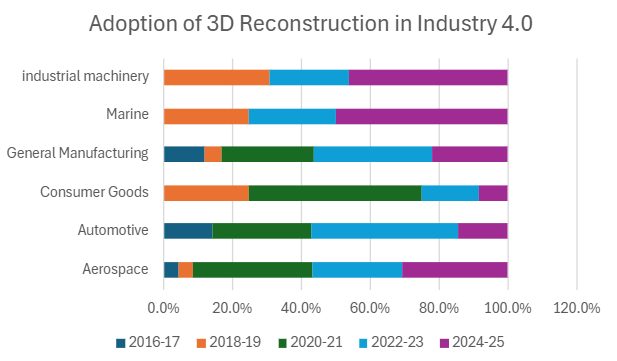}
        \caption{Adoption in Industry 4.0 between year 2016-2025.}
        \label{Adoption in Industry 4.0 between year 2016-2025}
\end{figure}

Fig. \ref{Adoption in Industry 4.0 between year 2016-2025} illustrates the increasing adoption of 3D reconstruction technology across six major industrial sectors. The chart highlights a significant acceleration in the technology's integration, aligning with Industry 4.0's focus on automation, connectivity, and real-time data analytics. By 2024–2025, nearly all sectors demonstrate a marked increase in adoption, indicating the transition of 3D reconstruction from an exploratory phase to a core component of industrial applications.\newline

In summary, the 3D reconstruction applications in manufacturing are classified into process-specific and industry-specific focuses. Process-specific applications include design and development, machining, process, assembly, and quality inspection. Quality inspection is the largest category, using 3D reconstruction to verify product quality and detect defects. Industry-specific applications span aerospace, automotive, consumer goods, general manufacturing, marine, and industrial machinery. For instance, the aerospace sector focuses on inspection, machining, and assembly, while automotive emphasizes quality control. The general manufacturing sector addresses design, assembly, and inspection with 3D reconstruction. The chart in Fig. \ref{Adoption in Industry 4.0 between year 2016-2025} shows an accelerating adoption of 3D reconstruction across all sectors, reflecting Industry 4.0's push towards automation and real-time analytics.\newline

\section{Challenges and Future Applications}
\label{sec:challenges_and_future}
\subsection{Challenges}

In sectors such as aerospace and automotive, sub-millimeter precision is essential; however, environmental disturbances—such as inconsistent lighting, vibrations, and occlusions—can introduce significant errors into the reconstruction pipeline.

In addictive manufacturing, Anmar et al. \cite{ammar_application_2019} emphasized that current 3D reconstruction techniques frequently encounter processing speed constraints. These limitations not only extend the overall manufacturing cycle but also hinder the effectiveness of real-time defect detection, which is particularly critical for applications that demand short production cycles.

Scalability also presents a major limitation. Applications like shipbuilding and large-scale aerospace assembly involve extensive and heterogeneous surfaces, which current techniques often struggle to reconstruct efficiently due to computational overhead and sensor coverage limitations.
The integration of AI and machine learning offers promising avenues for automating preprocessing and enhancing surface modeling. Nevertheless, the success of such models depends on access to large, diverse datasets and the ability to generalize across varying industrial contexts.
Furthermore, the adoption of 5G and IoT infrastructure is critical to support real-time 3D reconstruction in distributed manufacturing systems. These technologies enable fast, synchronized data transfer across devices, yet require robust cybersecurity and data management frameworks.
Hybrid systems that combine multiple sensor modalities—such as structured light, LiDAR, and depth cameras—are being explored to improve robustness and adaptability. However, real-time data fusion and cross-calibration remain active research challenges, especially in dynamic and unstructured environments.

One critical challenge is balancing trade-offs between sensor resolution, speed, and adaptability. For example, while fringe projection scanners deliver micron-level resolution, they struggle with glossy or transparent materials, which may require hybrid systems incorporating WLI or NeRF-based reflectance modeling. Moreover, integrating high-fidelity sensors like confocal microscopes into robotic workflows poses issues of mechanical complexity and data volume. Future directions may benefit from real-time multi-modal fusion frameworks where low-resolution depth maps are enhanced using learned priors from photometric or interferometric cues.

\subsection{Future Applications}
Future advancements will drive 3D reconstruction toward becoming a core component of intelligent manufacturing systems. One key application is real-time defect detection and predictive maintenance. By integrating AI-driven reconstruction with digital twins, manufacturers can identify anomalies early, reduce downtime, and enable data-driven maintenance strategies.
Robotic systems will increasingly rely on 3D reconstruction for adaptive machining and assembly. Reconstruction data will guide real-time trajectory planning and support collaboration between multiple robots in complex tasks, improving process flexibility and autonomy.
It has been shown that Gaussian Splatting can achieve significantly faster rendering and comparable visual fidelity compared to NeRF in certain scenarios, particularly for real-time applications \cite{kerbl2023gaussian} and it will play a pivotal role in digital twin development. These methods offer high-fidelity representations of manufacturing environments, enabling enhanced simulation, virtual testing, and process optimization.
Lastly, human-robot collaboration in smart factories will be elevated through continuous 3D environment mapping. With the ability to perceive and respond to dynamic changes in the workspace, robots will assist human operators more safely and effectively, facilitating cooperative workflows in increasingly autonomous production lines.

\section{CONCLUSIONS}
\label{sec:conclusion}
3D reconstruction has emerged as a transformative technology in manufacturing, enhancing design, machining, assembly, and quality inspection. It enables precise digitalization, improves production efficiency, and facilitates automation, aligning with Industry 4.0’s goals of connectivity and real-time analytics. Applications span across sectors, from aerospace to consumer goods, demonstrating its versatility in optimizing processes.

Despite its advantages, challenges persist. High computational costs, data processing complexities, and hardware limitations hinder seamless integration. Accuracy issues, especially in dynamic environments, require further advancements in AI-driven algorithms and sensor fusion. Standardization and interoperability also remain critical concerns.

Future developments will likely focus on AI-enhanced reconstruction, real-time processing, and cloud-based solutions for scalable implementation. Integration with digital twins and AR/VR could revolutionize predictive maintenance, simulation, and remote monitoring. As manufacturing shifts toward intelligent automation, 3D reconstruction will become increasingly indispensable in ensuring precision and efficiency. Continued research and industry collaboration will drive innovation, overcoming existing limitations and unlocking new possibilities.

In summary, while challenges remain, 3D reconstruction is poised to play a crucial role in shaping the future of smart manufacturing, bridging the gap between physical and digital environments for improved productivity and decision-making.\newline

\textbf{CRediT authorship contribution statement}\newline
\textbf{Chialoon Cheng }- Conceptualization, Supervision, Project administration, Investigation, Data curation, Writing - Original draft, Writing - review \& editing.
\textbf{Kaijun Liu} - Conceptualization, Formal Analysis, Visualization, Writing - review \& editing.
\textbf{Liu Zhiyang} - Conceptualization, Methodology, Writing – review \& editing.
\textbf{Marcelo Ang Jr} - Writing – review \& editing, Supervision.\newline

\textbf{Funding sources}\newline
This research did not receive any specific grant from funding agencies in the public, commercial, or not-for-profit sectors.

\clearpage
\bibliography{3d_recon_technique}
\bibliographystyle{IEEEtran}

\end{document}